\documentclass[twoside,11pt]{article}
\input epsf
\usepackage{modified_jair}
\usepackage{theapa}
\usepackage{amsmath}
\usepackage{float}
\usepackage{mathptmx}        
\usepackage{helvet}          
\usepackage{courier}         
\usepackage{type1cm}         
\usepackage{rotating}
\usepackage{graphicx}        
                             
\usepackage{multicol}        
\usepackage[bottom]{footmisc}
\usepackage{times}
\usepackage{latexsym}
\usepackage{url}
\usepackage{xspace}
\usepackage{arydshln}        

\usepackage{examples}
\exampleindent=\leftmargini
\newcommand{\pref}[1]{(\ref{#1})} 
\newenvironment{examps}[0]        
   {\begin{examples*}
    \setlength{\itemsep}{1pt}
    \setlength{\parskip}{0pt}
    \setlength{\parsep}{0pt}
    \small}
   {\end{examples*}}

\newcommand{\code}[1]{{\small \texttt{#1}}} 

\newcommand{\rte}{\textsc{rte}\xspace}

\newcommand{\svm}{\textsc{svm}\xspace}
\newcommand{\svms}{\textsc{svm}s\xspace}

\newcommand{\pos}{\textsc{pos}\xspace}

\newcommand{\msr}{\textsc{msr}\xspace}

\newcommand{\dirt}{\textsc{dirt}\xspace}
\newcommand{\ledir}{\textsc{ledir}\xspace}
\newcommand{\fm}{$F$-measure\xspace}

\newcommand{\smt}{\textsc{smt}\xspace}

\newcommand{\ibm}{\textsc{ibm}\xspace}
\newcommand{\basea}{\textsc{base$_1$}\xspace}
\newcommand{\baseb}{\textsc{base$_2$}\xspace}

\newcommand{\qa}{\textsc{qa}\xspace}

\newcommand{\tease}{\textsc{tease}\xspace}

\updateheading{May 30, 2010}
\ShortHeadings{A Survey of Paraphrasing and Textual Entailment Methods}{Androutsopoulos \& Malakasiotis}

\begin{document}
\firstpageno{1}
\title{A Survey of Paraphrasing and Textual Entailment Methods}
\author{\name Ion Androutsopoulos \email ion@aueb.gr \\
        \name Prodromos Malakasiotis \email rulller@aueb.gr \\
        \addr Department of Informatics\\
        Athens University of Economics and Business\\ 
        Patission 76, GR-104 34 Athens, Greece}
\maketitle

\begin{abstract}
Paraphrasing methods recognize, generate, or extract phrases, sentences, or longer natural language expressions that convey almost the same information. Textual entailment methods, on the other hand, recognize, generate, or extract pairs of natural language expressions, such that a human who reads (and trusts) the first element of a pair would most likely infer that the other element is also true. Paraphrasing can be seen as bidirectional textual entailment and methods from the two areas are often 
similar. Both kinds of methods are useful,
at least in principle, in a wide range of natural language processing applications, including question answering, summarization, text generation, and machine translation. We summarize key ideas from the two areas by considering in turn recognition, generation, and extraction methods, also pointing to prominent articles and resources.  
\end{abstract}

\section{Introduction} \label{sec:introduction}

This article is a survey of computational methods for paraphrasing and textual entailment. Paraphrasing methods recognize, generate, or extract (e.g., from corpora) paraphrases, meaning phrases, sentences, or longer texts that convey the same, or almost the same information. For example, 
\pref{bridge1} and \pref{bridge2} are paraphrases. Most people would also accept 
\pref{bridge3} as a paraphrase of 
\pref{bridge1} and \pref{bridge2}, though it could be argued that in 
\pref{bridge3} the construction of the bridge has not necessarily been completed, unlike
\pref{bridge1} and \pref{bridge2}.\footnote{
Readers familiar with tense and aspect theories will have recognized that  \pref{bridge1}--\pref{bridge3} involve an ``accomplishment'' of Vendler's  \citeyear{Vendler1967} taxonomy. The accomplishment's completion point is not necessarily reached in \pref{bridge3}, unlike \pref{bridge1}--\pref{bridge2}.}
Such fine distinctions, however, are usually 
ignored in paraphrasing and textual entailment work, which is why we say that paraphrases may convey \emph{almost} the same information. 

\begin{examps}
\item Wonderworks Ltd.\ constructed the new bridge.\label{bridge1}
\item The new bridge was constructed by Wonderworks Ltd.\label{bridge2}
\item Wonderworks Ltd.\ is the constructor of the new bridge.\label{bridge3}
\end{examps} 

Paraphrasing methods may also operate on \emph{templates} of natural language expressions,  like \pref{pattern1}--\pref{pattern3}; here the slots $X$ and $Y$ can be filled in with arbitrary noun phrases. Templates specified at the syntactic or semantic level may also be used, where the slot fillers may be required to have particular syntactic relations (e.g., verb-object) to other words or constituents, or to satisfy semantic constraints (e.g., requiring $Y$ to denote a book). 

\begin{examps}
\item $X$ wrote $Y$.\label{pattern1}
\item $Y$ was written by $X$.\label{pattern2}
\item $X$ is the writer of $Y$.\label{pattern3}
\end{examps} 

Textual entailment methods, on the other hand, recognize, generate, or extract pairs $\left<T, H\right>$ of natural language expressions, such that a human who reads (and trusts) $T$ would infer that $H$ is most likely also true \cite{Dagan2006}. For example, \pref{T1} textually entails \pref{H1}, but \pref{T2} does not textually entail \pref{H2}.\footnote{Simplified examples from \rte-2 \cite{Bar-Haim2006}.} 

\begin{examps}
\item The drugs that slow down Alzheimer's disease work best the earlier you administer them.\label{T1}
\item Alzheimer's disease can be slowed down using drugs.\label{H1}
\item Drew Walker, Tayside's public health director, said: ``It is important to stress that this is not a confirmed case of rabies.''\label{T2}
\item A case of rabies was confirmed.\label{H2}
\end{examps}

As in paraphrasing, textual entailment methods may operate on templates. For example, in a discourse about painters, composers, and their work, \pref{pattern5} textually entails \pref{pattern6}, for any noun phrases $X$ and $Y$. However, \pref{pattern6} does not textually entail \pref{pattern5}, when $Y$ denotes a symphony composed by $X$. If we require textual entailment between templates to hold for all possible slot fillers, then \pref{pattern5} textually entails \pref{pattern6} in our example's discourse, but the reverse does not hold.

\begin{examps}
\item $X$ painted $Y$.\label{pattern5}
\item $Y$ is the work of $X$.\label{pattern6} 
\end{examps}

In general, we cannot judge if two natural language expressions are paraphrases or a correct textual entailment pair without selecting particular \emph{readings} of the expressions, among those that may be possible due to multiple word senses, syntactic ambiguities etc. For example, \pref{bank1} textually entails \pref{bank2} with the financial sense of ``bank'',  but not when \pref{bank1} refers to the bank of a river. 

\begin{examps}
\item A bomb exploded near the French bank.\label{bank1}
\item A bomb exploded near a building.\label{bank2}
\end{examps}

One possibility, then, is to examine the language expressions (or templates) only in particular contexts that make their intended readings clear. Alternatively, we may want to treat as correct any textual entailment pair $\left<T, H\right>$ for which there are \emph{possible} readings of $T$ and $H$, such that a  human who reads $T$ would infer that $H$ is most likely also true; then, if a system reports that \pref{bank1} textually entails \pref{bank2}, its response is to be counted as correct, regardless of the intended sense of ``bank''. Similarly, paraphrases would have \emph{possible} readings conveying almost the same information. 

The lexical substitution task of \textsc{semeval} \cite{McCarthy2009}, where systems are required to find 
an appropriate substitute for a particular word in the context of a given sentence, can be seen as a special case of paraphrasing or textual entailment, restricted to pairs of words. \textsc{semeval}'s task, however, includes the requirement that it must be possible to use the two words (original and replacement) in exactly the same context. In a similar manner, one could adopt a stricter definition of paraphrases, which would require them not only to have the same (or almost the same) meaning, but also to be expressions that can be used interchangeably 
in grammatical sentences. In that case, although \pref{Edison1} and \pref{Edison2} are paraphrases, their underlined parts are not, because they cannot be swapped in the two sentences; the resulting sentences would be ungrammatical. 

\begin{examps}
\item \underline{Edison invented the light bulb in 1879}, providing a long lasting source of light.\label{Edison1}
\item \underline{Edison's invention of the light bulb in 1879} provided a long lasting source of light.\label{Edison2}
\end{examps}

\noindent A similar stricter definition of textual entailment would impose the additional requirement that $H$ and $T$ can replace each other in grammatical sentences.

\subsection{Possible Applications of Paraphrasing and Textual Entailment Methods}
\label{applications}

The natural language expressions that paraphrasing and textual entailment methods consider are not always statements. 
In fact, many of these methods were developed having question answering (\qa) systems in mind. In \qa systems for document collections \cite{Voorhees2001,Pasca2003,Harabagiu2003b,Molla2007}, a question may be phrased differently than in a document that contains the answer, and taking such variations into account can improve system performance significantly 
\cite{Harabagiu2003,Duboue2006,Harabagiu2006,Riezler2007}. 
For example, a \qa system may retrieve relevant documents or passages, using the input question as a query to an information retrieval or Web search engine \cite{BaezaYates1999,Manning2008}, and then check if any of the retrieved texts textually entails a candidate answer \cite{Moldovan2001,Duclaye2003}.\footnote{
Culicover \citeyear{Culicover1968} discussed different types of paraphrasing and entailment, and proposed the earliest computational treatment of paraphrasing and textual entailment that we are aware of, with the goal of retrieving passages of texts that answer natural language queries. We thank one of the anonymous reviewers for pointing us to Culicover's work.}
If the input question is \pref{question4} and the search engine returns passage \pref{snippet1}, the system may check if \pref{snippet1} textually entails any of the candidate answers of \pref{candidate1}, where we have replaced the interrogative ``who'' of \pref{question4} with all the expressions of \pref{snippet1} that a named entity recognizer \cite{Bikel1999,Sekine2009} would ideally have recognized as person names.\footnote{Passage \pref{snippet1} is based on Wikipedia's page for Doryphoros.}

\begin{examps}
\item Who sculpted the Doryphoros?\label{question4}
\item The Doryphoros is one of the best known Greek sculptures of the classical era in Western Art. The Greek sculptor \emph{Polykleitos} designed this work as an example of the ``canon'' or ``rule'', showing the perfectly harmonious and balanced proportions of the human body in the sculpted form. The sculpture was known through the Roman marble replica found in Herculaneum and conserved in the Naples National Archaeological Museum, but, according to \emph{Francis Haskell} and \emph{Nicholas Penny}, early connoisseurs passed it by in the royal Bourbon collection at Naples without notable comment.\label{snippet1}
\item Polykleitos/Francis Haskell/Nicholas Penny sculpted the Doryphoros.\label{candidate1}
\end{examps}

The input question may also be paraphrased, to allow more, potentially relevant passages to be obtained. 
Question paraphrasing is also useful when mapping user questions to lists of frequently asked questions (\textsc{faq}s) that are accompanied by their answers \cite{Tomuro2003};
and natural language interfaces to databases 
often generate question paraphrases to allow users to understand if their queries have been understood \cite{McKeown1983,Androutsopoulos1995}. 

Paraphrasing and textual entailment methods are also useful in several other natural language processing applications. In text summarization \cite{Mani2001,Hovy2003}, 
for example, an important processing stage is typically sentence extraction, which identifies the most important sentences of the texts to be summarized. During that stage, especially when generating a single summary from several documents \cite{Barzilay2005},
it is important to avoid selecting sentences 
(e.g., from different news articles about the same event) that convey the 
same information (paraphrases) as other sentences that have already been selected, or sentences whose information follows from other already selected sentences (textual entailment).

Sentence compression \cite{Knight2002,McDonald2006,CohnLapata2008,ClarkeLapata2008,CohnLapata2009,Galanis2010},
often also a processing stage of text summarization, can be seen as a special case of
sentence paraphrasing, as suggested by Zhao et al.\ \citeyear{Zhao2009}, with the additional 
constraint that the resulting sentence
must be shorter than the original one and still grammatical; 
for example, a sentence matching \pref{pattern2} or \pref{pattern3} could be shortened by converting it to a paraphrase of the form of \pref{pattern1}. 
Most sentence compression work, however, allows less important information of the original sentence to be discarded. Hence, the resulting sentence is entailed by, it is not necessarily a paraphrase of the original one.
In the following example, \pref{compressOut} is a compressed form of \pref{compressIn} produced by a human.\footnote{Example from Clarke et al.'s paper, ``Written News Compression Corpus \cite{ClarkeLapata2008}; see Appendix \ref{resources}.} 

\begin{examps}
\item Mother Catherine, 82, the mother superior, will attend the hearing on Friday, he said. \label{compressIn}
\item Mother Catherine, 82, the mother superior, will attend. \label{compressOut}
\end{examps}

\noindent When the compressed sentence is not necessarily a paraphrase of the original one, we may first produce (grammatical) candidate compressions that are textually entailed by the original sentence; hence, a mechanism to generate textually entailed sentences is useful. Additional mechanisms are needed, however, to rank the candidates depending on the space they save and the degree to which they maintain important information; we do not discuss  additional mechanisms of this kind. 

Information extraction systems \cite{Grishman2003,Moens2006} often rely on manually or automatically crafted patterns \cite{Muslea1999} to  
locate text snippets that report particular types of events and to identify the entities involved; for example, patterns like \pref{ie1}--\pref{ie3}, or similar patterns operating on syntax trees, possibly with additional semantic constraints, might be used to locate snippets referring to bombing incidents and identify their targets.
Paraphrasing or textual entailment methods can be used to generate additional semantically equivalent extraction patterns (in the case of paraphrasing) or patterns that textually entail the original ones \cite{Shinyama2002}.  

\begin{examps}
\item $X$ was bombed \label{ie1}
\item bomb exploded near $X$ \label{ie2}
\item explosion destroyed $X$ \label{ie3}
\end{examps}

In machine translation \cite{Koehn2009}, ideas from paraphrasing and textual entailment research have been embedded in measures and processes that automatically evaluate machine-generated translations against human-authored ones that may use different phrasings \cite{Lepage2005,Zhou2006,Kauchak2006,Pado2009}; we return to this issue in following sections.
Paraphrasing methods have also been used to automatically generate additional reference translations from human-authored ones when training machine translation systems \cite{Madnani2007}.
Finally, paraphrasing and textual entailment methods have been employed to allow machine translation systems to cope with source language words and longer phrases that have not been encountered in training corpora \cite{Zhang2005b,CallisonBurch2006,Marton2009,Mirkin2009}. 
To use an example of Mirkin et al.\ \citeyear{Mirkin2009}, a phrase-based machine translation system that has never encountered the expression ``file a lawsuit'' during its training, but which knows that pattern \pref{cisco3} textually entails \pref{cisco4}, may be able to produce a more acceptable translation by converting \pref{cisco1} to \pref{cisco2}, and then translating \pref{cisco2}. Some information would be lost in the translation, because \pref{cisco2} is not a paraphrase of \pref{cisco1}, but the translation may still be preferable to the outcome of translating directly \pref{cisco1}.

\begin{examps}
\item $X$ filed a lawsuit against $Y$ for $Z$.\label{cisco3}
\item $X$ accused $Y$ of $Z$.\label{cisco4}
\item Cisco filed a lawsuit against Apple for patent violation.\label{cisco1}
\item Cisco accused Apple of patent violation.\label{cisco2}
\end{examps}

In natural language generation \cite{Reiter2000,Bateman2003}, 
for example when producing texts describing the entities of a formal ontology \cite{ODonnell2001,Androutsopoulos2007}, paraphrasing can be used to avoid repeating the same 
phrasings (e.g., when expressing properties of similar entities), or to produce alternative expressions that improve text coherence, 
adhere to writing style (e.g., avoid passives), or satisfy other constraints \cite{Power2005}.
Among other possible applications, paraphrasing and textual entailment methods can 
be employed to simplify texts, for example by replacing specialized (e.g., medical) terms with expressions non-experts can understand \cite{Elhadad2007,Deleger2009}, and to automatically score student answers against reference answers \cite{Nielsen2009}.

\subsection{The Relation of Paraphrasing and Textual Entailment to Logical Entailment}

If we represent the meanings of natural language expressions by 
logical formulae, for example in first-order predicate logic, we may think of textual entailment and paraphrasing in terms of logical entailment ($\models$). If the logical meaning representations of $T$ and $H$ are $\phi_T$ and $\phi_H$, then $\left<T, H\right>$ is a correct textual entailment pair if and only if $(\phi_T \wedge B) \models \phi_H$; $B$ is a knowledge base, for simplicity assumed here to have the form of a single conjunctive formula, which contains meaning postulates \cite{carnap1952} and other knowledge assumed to be shared by all language users.\footnote{Zaenen et al.\ \citeyear{Zaenen2005} provide examples showing that linguistic and world knowledge cannot often be separated.} Let us consider the example below, where logical terms starting with capital letters are constants; we assume that different word senses would give rise to different predicate symbols. Let us also assume that $B$ contains only $\psi$. Then $(\phi_T \wedge \psi) \models \phi_H$  holds, i.e., $\phi_H$ is true for any interpretation (e.g., model-theoretic) of constants, predicate names and other domain-dependent atomic symbols, for which $\phi_T$ and $\psi$ both hold. A sound and complete automated reasoner (e.g., based on resolution, in the case of first-order predicate logic) could be used to confirm that the logical entailment holds. Hence, $T$ textually entails $H$, assuming again that the meaning postulate $\psi$ is available. 
The reverse, however, does not hold, i.e., $(\phi_H \wedge \psi) \not \models \phi_T$; the implication ($\Rightarrow$) of $\psi$ would have to be made bidirectional ($\Leftrightarrow$) for the reverse to hold. 

{\small
\begin{eqnarray*}
T & : & \textrm{Leonardo da Vinci painted the Mona Lisa.}\\
\phi_T & :&  \textit{isPainterOf}(\textit{DaVinci}, \textit{MonaLisa})\\
H & : & \textrm{Mona Lisa is the work of Leonardo da Vinci.}\\
\phi_H & : & \textit{isWorkOf}(\textit{MonaLisa}, \textit{DaVinci})\\
\psi & : & \forall x \, \forall y \; \textit{isPainterOf}(x, y) \Rightarrow \textit{isWorkOf}(y, x)
\end{eqnarray*}
}

Similarly, if the logical meaning representations of $T_1$ and $T_2$ are $\phi_1$ and $\phi_2$, then $T_1$ is a paraphrase of $T_2$ iff $(\phi_1 \wedge B) \models \phi_2$ and $(\phi_2 \wedge B) \models \phi_1$, where again $B$ contains meaning postulates and common sense knowledge. Ideally, sentences like 
\pref{bridge1}--\pref{bridge3} would be represented by the same formula, making it clear that they are paraphrases, regardless of the contents of $B$. Otherwise, it may sometimes be unclear if $T_1$ and $T_2$ should be considered paraphrases, because it may be unclear if some knowledge should be considered part of $B$. 

Since natural language expressions are often ambiguous, especially out of context, we may again want to adopt looser definitions, so that $T$ textually entails $H$ iff there are \emph{possible} readings of $T$ and $H$, represented by $\phi_H$ and $\phi_T$, such that $(\phi_T \wedge B) \models \phi_H$, and similarly for paraphrases. Thinking of textual entailment and paraphrasing in terms of logical entailment allows us to borrow notions and methods from logic. Indeed, some paraphrasing and textual entailment recognition methods map natural language expressions to logical formulae, and then examine if logical entailments hold. This is not, however, the only possible approach. Many other, if not most, methods currently operate on surface strings or syntactic representations, without mapping natural language expressions to formal meaning representations. Note, also, that in methods that map natural language to logical formulae, it is important to work with a form of logic that provides adequate support for logical entailment checks; full first-order predicate logic may be inappropriate, as it is semi-decidable.

To apply our logic-based definition of textual entailment,
which was formulated for statements, to questions, let us use identical fresh constants (in effect, Skolem constants) across questions to represent the unknown entities the questions ask for; we mark such constants with question marks as subscripts, but in logical entailment checks they can be treated as ordinary constants.
In the following example, the user asks $H$, and the system generates $T$.
Assuming that the meaning postulate $\psi$ is available in $B$, $(\phi_T \wedge B) \models \phi_H$, i.e., for any interpretation of the predicate symbols and constants, if $(\phi_T \wedge B)$ is true, then $\phi_H$ is necessarily also true. Hence, $T$ textually entails $H$. 
In practice, this means that if the system manages to find an answer to $T$, perhaps because $T$'s phrasing is closer to a sentence in a document collection, the same answer can be used to respond to $H$. 

{\small
\begin{eqnarray*}
T (generated) & : & \textrm{Who painted the Mona Lisa?} \\
\phi_T & : & \textit{isAgent}(W_?) \wedge \textit{isPainterOf}(W_?, \textit{MonaLisa})\\H (asked) & : & \textrm{Whose work is the Mona Lisa?} \\
\phi_H & : & \textit{isAgent}(W_?) \wedge \textit{isWorkOf}(\textit{MonaLisa}, W_?) \\
\psi & : & \forall x \, \forall y \; \textit{isPainterOf}(x, y) \Rightarrow \textit{isWorkOf}(y, x)
\end{eqnarray*}
}

A logic-based definition of question paraphrases can be formulated in a similar manner, as bidirectional logical entailment. 
Note also that logic-based paraphrasing and textual entailment methods may actually represent interrogatives as free variables, instead of fresh constants, and they may rely on unification to obtain their values \cite{Moldovan2001,rinaldi03}.

\subsection{A Classification of Paraphrasing and Textual Entailment Methods}

There have been six workshops on paraphrasing and/or textual entailment \cite{Sato2001,Inui2003,Dolan2005,Drass2005,Sekine2007,CallisonBurch2009} in recent years.\footnote{The proceedings of the five more recent workshops are available in the \textsc{acl} Anthology.} The Recognizing Textual Entailment (\textsc{rte}) challenges \cite{Dagan2006,Bar-Haim2006,Giampiccolo2007,Giampiccolo2008}, currently in their fifth year, provide additional significant thrust.\footnote{The \textsc{rte} challenges were initially organized by the European \textsc{pascal} Network of Excellence, and subsequently as part of \textsc{nist}'s Text Analysis Conference.} 
Consequently, there is a large number of published articles, proposed methods, and  resources related to paraphrasing and textual entailment.\footnote{A textual entailment portal has been established, as part of \textsc{acl}'s wiki, to help organize all relevant material.}  
A special issue on textual entailment was also recently published, and its editorial provides a brief overview of textual entailment methods \cite{Dagan2009}.\footnote{The slides of Dagan, Roth, and Zazotto's \textsc{acl} 2007 tutorial on textual entailment are also 
publicly available.} To the best of our knowledge, however, 
the present article is the first
extensive survey of 
paraphrasing and textual entailment. 

To provide a clearer view of the different goals and assumptions of the methods that have been proposed, we classify them along two dimensions: whether they are \emph{paraphrasing} or \emph{textual entailment} methods; and whether they perform \emph{recognition}, \emph{generation}, or \emph{extraction} of paraphrases or textual entailment pairs. These
distinctions are not always clear in the literature, especially the distinctions along the second dimension, which we explain below. It is also possible to classify methods along other dimensions, for example depending on whether they operate on language expressions or templates;
or whether they operate on phrases, sentences or longer texts.

The main input to a paraphrase or textual entailment \emph{recognizer} is a pair of language expressions (or templates), possibly in particular contexts. The output is a judgement, possibly probabilistic, indicating whether or not the members of the input pair are paraphrases or a correct textual entailment pair; the judgements must agree as much as possible with those of humans. On the other hand, the main input to a paraphrase or textual entailment \emph{generator} is a single language expression (or template) at a time, possibly in a particular context. The output is a set of paraphrases of the input, or a set of language expressions that entail or are entailed by the input; the output set must be as large as possible, but including as few errors as possible. In contrast, no particular language expressions or templates are provided to a paraphrase or textual entailment \emph{extractor}. The main input in this case is a corpus, for example a monolingual corpus of parallel or comparable texts, such as different English translations of the same French novel, or clusters of multiple monolingual news articles, with the articles in each cluster reporting the same event. The system outputs pairs of paraphrases (possibly templates), or pairs of language expressions (or templates) that constitute correct textual entailment pairs, based on the evidence of the corpus; the goal is again to produce as many output pairs as possible, with as few errors as possible. Note that the boundaries between recognizers, generators, and extractors may not always be clear. For example, a paraphrase generator may invoke a paraphrase recognizer to filter out erroneous candidate paraphrases;
and a recognizer or a generator may consult a collection of template pairs produced by an extractor. 

We note that articles reporting actual applications of paraphrasing and textual entailment methods to larger systems (e.g., for \qa, information extraction, machine translation, as discussed in Section \ref{applications}) are currently relatively few, compared to the number of articles that propose new paraphrasing and textual entailment methods or that test them in vitro, despite the fact that articles of the second kind very often point to possible applications of the methods they propose. The relatively small number of application articles may be an indicator that paraphrasing and textual entailment methods are not used extensively in larger systems yet. We believe that this may be due to at least two reasons. First, the efficiency of the methods needs to be improved, which may require combining recognition, generation, and extraction methods, for example to iteratively produce more training data; we return to this point in following sections. Second, the literature on paraphrasing and textual entailment is vast, which makes it difficult for researchers working on larger systems to assimilate its key concepts and identify suitable methods. We hope that this article will help address the second problem, while also acting as an introduction that may help new researchers improve paraphrasing and textual entailment methods further. 

In Sections \ref{sec:recognition}, \ref{sec:generation}, and \ref{sec:extraction} below we consider in turn recognition, generation, and extraction methods for both paraphrasing and textual entailment. In each of the three sections, we attempt to identify and explain prominent ideas, pointing also to relevant articles and resources. In Section \ref{section:conclusions}, we conclude and discuss some possible directions for future research. The \textsc{url}s of all publicly available resources that we mention are listed in appendix \ref{resources}.

\section{Paraphrase and Textual Entailment Recognition} \label{sec:recognition}

Paraphrase and textual entailment recognizers judge whether or not two given language expressions (or templates) constitute paraphrases or a correct textual entailment pair. 
Different methods may operate at different levels of representation of the input expressions; for example, they may treat the input expressions simply as surface strings, they may operate on syntactic or semantic representations of the input expressions, or on representations combining information from different levels.

\subsection{Logic-based Approaches to Recognition}

One possibility is to map the language expressions to logical meaning representations, and then rely on logical entailment checks, possibly by invoking theorem provers
\cite{rinaldi03,Bos2005,Tatu2005,tatu-moldovan:2007:WTEP}. In the case of textual entailment, this involves generating pairs of formulae $\left<\phi_T,\phi_H\right>$ for $T$ and $H$ (or their possible readings), and then checking if $(\phi_T \wedge B) \models \phi_H$, where $B$ contains meaning postulates and common sense knowledge, as already discussed. In practice, however, it may be very difficult to formulate a reasonably complete $B$. A partial solution to this problem is to obtain common sense knowledge from resources like WordNet \cite{fellbaum98} or Extended WordNet \cite{Moldovan2001}. The latter also includes logical meaning representations extracted from WordNet's glosses. For example, since ``assassinate'' is a hyponym (more specific sense) of ``kill'' in WordNet, an axiom like the following can be added to $B$  \cite{Moldovan2001,Bos2005,tatu-moldovan:2007:WTEP}. 
\[
\forall x \, \forall y \; \textit{assassinate}(x, y) \Rightarrow \textit{kill}(x, y)
\]

Additional axioms can be obtained from FrameNet's frames \cite{Baker1998,LonnekerRodman2009}, as discussed for example by Tatu et al.\ \citeyear{Tatu2005}, or similar resources. Roughly speaking, a frame is the representation of a prototypical situation (e.g., a purchase), which also identifies the situation's main roles (e.g., the buyer, the entity  bought), the types of entities (e.g., person) that can play these roles, and possibly relations (e.g., causation, inheritance) to other prototypical situations (other frames). VerbNet \cite{KipperSchuler2005} also specifies, among other information, semantic frames for English verbs. On-line encyclopedias  have also been used to obtain background knowledge by extracting particular types of information (e.g., is-a relationships) from their articles \cite{iftene-balahurdobrescu:2007:WTEP}. 

Another approach is to use no particular $B$ 
(meaning postulates and common sense knowledge), and measure how difficult it is to satisfy both $\phi_T$ and $\phi_H$, in the case of textual entailment recognition, compared to satisfying $\phi_T$ on its own. A possible measure is the difference of the size of the minimum model that satisfies both $\phi_T$ and $\phi_H$, compared to the size of the minimum model that satisfies $\phi_T$ on its own \cite{Bos2005}; intuitively, a model is an assignment of entities, relations etc.\ to terms, predicate names, and other domain-dependent atomic symbols. The greater this difference the more knowledge is required in $B$ for $(\phi_T \wedge B) \models \phi_H$ to hold, and the more difficult it becomes for speakers to accept that $T$ textually entails $H$. Similar bidirectional logical entailment checks can be used to recognize paraphrases 
\cite{rinaldi03}.

\subsection{Recognition Approaches that Use Vector Space Models of Semantics}
\label{vector_semantics}

An alternative to using logical meaning representations is to start by mapping each word of the input language expressions to a vector that shows how strongly the word cooccurs with particular other words in corpora \cite{Lin1998}, possibly also taking into account syntactic information, for example requiring that the cooccurring words participate in particular syntactic dependencies \cite{Pado2007}. A compositional vector-based meaning representation theory can then be used to combine the vectors of single words, eventually mapping each one of the two input expressions to a single vector that attempts to capture its meaning; in the simplest case, the vector of each expression could be the sum or product of the vectors of its words, but more elaborate approaches have also been proposed \cite{Mitchell2008,Erk2009,Clarke2009}. 
Paraphrases can then be detected by measuring the distance of the vectors of the two input  expressions, for example by computing their cosine similarity. 
See also the work of Turney and Pantel \citeyear{Turney2010} for a survey of vector space models of semantics. 

Recognition approaches based on vector space models of semantics appear to have been explored much less than other approaches discussed in this article, and mostly in paraphrase recognition \cite{Erk2009}. They could also be used in textual entailment recognition, however, by checking if the vector of $H$ is particularly close to that of a part (e.g., phrase or sentence) of $T$. Intuitively, this would check if what $H$ says is included in what $T$ says, though we must be careful with negations and other expressions that do not preserve truth values 
\cite{Zaenen2005,MacCartney2009}, as in \pref{nonmonotonic:1}--\pref{nonmonotonic:2}. We return to the idea of matching $H$ to a part of $T$ below.
\begin{examps}
\item $T$: He \emph{denied that} BigCo bought SmallCo.\label{nonmonotonic:1}
\item $H$: BigCo bought SmallCo.\label{nonmonotonic:2}
\end{examps}

\subsection{Recognition Approaches Based on Surface String Similarity}
\label{recognition_string}

Several paraphrase recognition methods operate directly on the input surface strings, possibly after applying some pre-processing, such as part-of-speech (\pos) tagging or named-entity recognition, but without computing more elaborate syntactic or semantic representations. For example, they may compute the string edit distance \cite{Levenshtein1966} of the two input strings, the number of their common words, or combinations of several string similarity measures \cite{malakasiotis-androutsopoulos:2007:WTEP}, including measures originating from machine translation evaluation \cite{Finch2005,Perez2005,Zhang2005,Wan2006}. The latter have been developed to automatically compare machine-generated translations against human-authored reference translations. A well known measure is \textsc{bleu} \cite{Papineni2002,Zhou2006}, which roughly speaking examines the percentage of word $n$-grams (sequences of consecutive words) of the machine-generated translations that also occur in the reference translations,
and takes the geometric average of the percentages obtained for different values of $n$.
Although such $n$-gram based measures have been criticised in machine translation evaluation \cite{CallisonBurch2006b}, for example because they are unaware of synonyms and longer paraphrases, they can be combined with other measures to build paraphrase (and textual entailment) recognizers \cite{Zhou2006,Kauchak2006,Pado2009}, which may help address the 
problems of automated machine translation evaluation. 

In textual entailment recognition, one of the input language expressions ($T$) is often much longer than the other one ($H$). If a part of 
$T$'s surface string is very similar to 
$H$'s, this is an indication that $H$ may be entailed by $T$. This is illustrated in \pref{deGaulle1}--\pref{deGaulle2}, where $H$ is included verbatim in $T$.\footnote{Example from the dataset of \rte-3 \cite{Giampiccolo2007}.} Note, however, that the 
surface string similarity (e.g., measured by string edit distance) between $H$ and the entire $T$ of this example is low, because of the different lengths of $T$ and $H$. 

\begin{examps}
\item $T$: \underline{Charles de Gaulle died in 1970} at the age of eighty. He was
thus fifty years old when, as an unknown officer recently promoted
to the rank of brigadier general, he made his famous
broadcast from London rejecting the capitulation of France to the
Nazis after the debacle of May-June 1940.\label{deGaulle1}

\item $H$: Charles de Gaulle died in 1970.\label{deGaulle2}
\end{examps}

\noindent Comparing $H$ to a sliding window of $T$'s surface string of the same size as $H$ (in our example, six consecutive words of $T$) and keeping the largest similarity score between the sliding window and $H$ may provide a better indication of whether $T$ entails $H$ or not \cite{Malakasiotis2009}. In many correct textual entailment pairs,
however, 
using a single sliding window of a fixed length may still be inadequate, because $H$ may correspond to several non-continuous parts of $T$; in \pref{quebec1}--\pref{quebec2}, for example, $H$ corresponds to the three underlined parts of $T$.\footnote{Modified example from the dataset of the \rte-3  \cite{Giampiccolo2007}.} 

\begin{examps}
\item $T$: \underline{The Gaspe}, also known as la Gaspesie in French, \underline{is a} North American \underline{peninsula} on the south shore of the Saint Lawrence River, \underline{in Quebec}.\label{quebec1}

\item $H$: The Gaspe is a peninsula in Quebec.\label{quebec2}
\end{examps}

One possible solution is to attempt to align the words (or phrases) of $H$ to those of $T$, and consider $T$--$H$ a correct textual entailment pair if a sufficiently good alignment is found, in the simplest case if a large percentage of $T$'s words are aligned to words of $H$. Another approach would be to use a 
window of variable length;
the window could be, for example, the shortest span of $T$ that contains all of $T$'s words that are aligned to words of $H$ \cite{Burchardt2009}. 
In any case, we need to be careful with negations and other expressions that do not preserve truth values, as already mentioned. Note, 
also, that although effective word alignment methods have been developed in statistical machine translation \cite{brown1993,Vogel1996,Och2003}, they often perform poorly on textual entailment pairs, because 
$T$ and $H$ are often of very different lengths, they do not necessarily convey the same information, and textual entailment training datasets are much smaller than those used in machine translation; see MacCartney et al.'s \citeyear{MacCartney2008b} work for further related discussion and a word alignment method developed especially for textual entailment pairs.\footnote{Cohn et al.\ \citeyear{Cohn2008} discuss how a 
publicly available corpus with manually word-aligned paraphrases was constructed. 
Other word-aligned paraphrasing or textual entailment datasets can be found at the \textsc{acl} Textual Entailment Portal.}

\subsection{Recognition Approaches Based on Syntactic Similarity}

\begin{figure}
\begin{centering}
\includegraphics[width=0.6\textwidth]{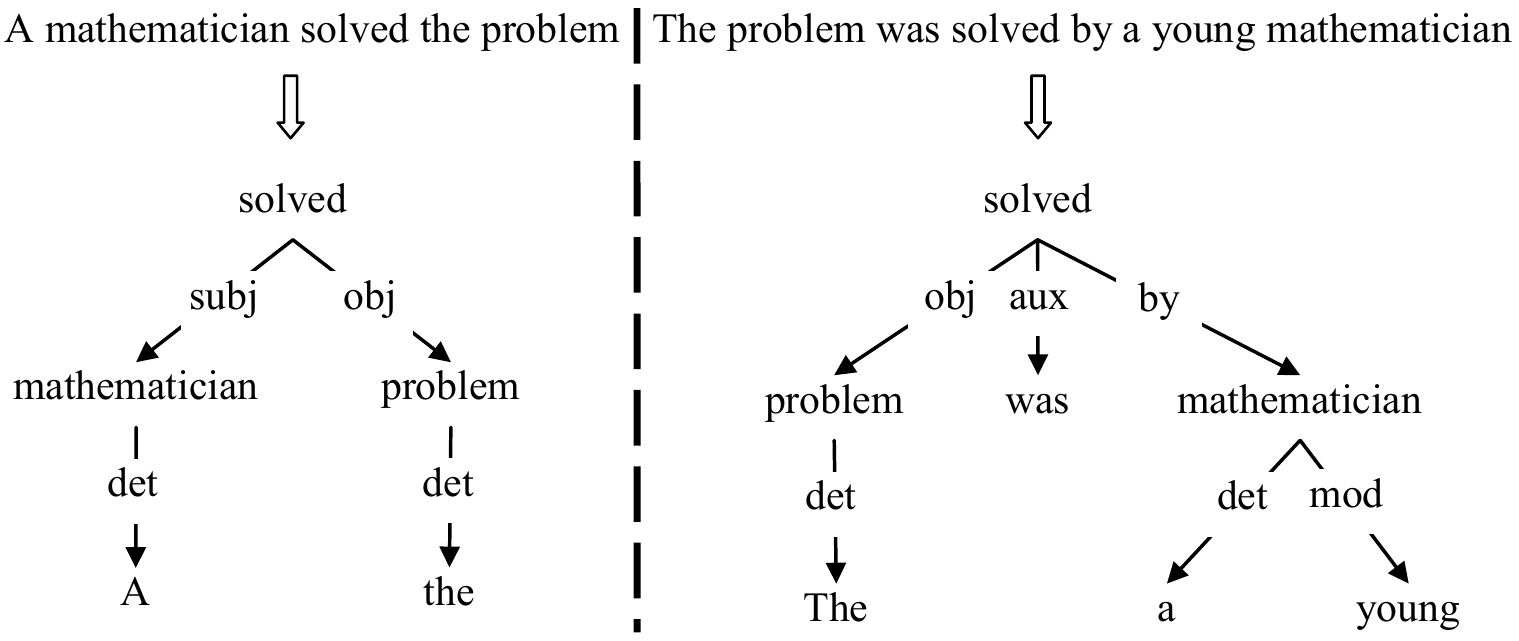}
\caption{Two sentences that are very similar when viewed at the level of dependency trees.} 
\label{fig:dependency_graphs}
\end{centering}
\end{figure}

Another common approach is to work at the syntax level. Dependency grammar parsers \cite{Melcuk1987,Kubler2009} 
are popular in paraphrasing and textual entailment research, as in other natural language processing areas recently. Instead of showing hierarchically the syntactic constituents (e.g., noun phrases, verb phrases) of a sentence, the output of a dependency grammar parser is a graph (usually a tree) whose nodes are the words of the sentence and whose (labeled) edges correspond to syntactic dependencies between words, for example the dependency between a verb and the head noun of its subject noun phrase, or the dependency between a noun and an adjective that modifies it. Figure \ref{fig:dependency_graphs} shows the dependency trees of two sentences. The exact form of the trees and the edge labels would differ, depending on the parser; for simplicity, we show prepositions as edges. If we ignore word order and the auxiliary ``was'' of the passive (right) sentence, and if we take into account that the \code{by} edge of the passive sentence corresponds to the \code{subj} edge of the active (left) one, the only difference is the extra adjective of the passive sentence. Hence, it is easy to figure out from the dependency trees that the two sentences have very similar meanings, despite their differences in word order. Strictly speaking, the right sentence textually entails the left one, not the reverse, because of the word ``young'' in the right sentence.

Some paraphrase recognizers simply count the common edges of the dependency trees of the input expressions \cite{Wan2006,Malakasiotis2009} or use other tree similarity measures. 
A large similarity 
score (e.g., above a threshold) indicates that the input expressions may be paraphrases. Tree edit distance \cite{Selkow1977,Tai1979,Zhang1989}
is 
another example of a similarity measure that can be applied to 
dependency or other parse trees; it computes the sequence of operator applications (e.g., add, replace, or remove a node or edge) with the minimum cost that turns one tree into the other.\footnote{\textsc{edits}, a suite to recognize textual entailment by computing edit distances, is publicly available.} To obtain more accurate predictions, it is important to devise an appropriate inventory of operators and assign appropriate costs to the operators during a training stage \cite{Kouylekov2005,Mehdad2009,Harmeling2009}.
For example, replacing a noun with one of its synonyms should be less costly than replacing it with an unrelated word; and removing a dependency between a verb and an adverb should perhaps be less costly than removing a dependency between a verb and the head noun of its subject or object. 

\begin{figure}
\begin{centering}
\includegraphics[width=0.7\textwidth]{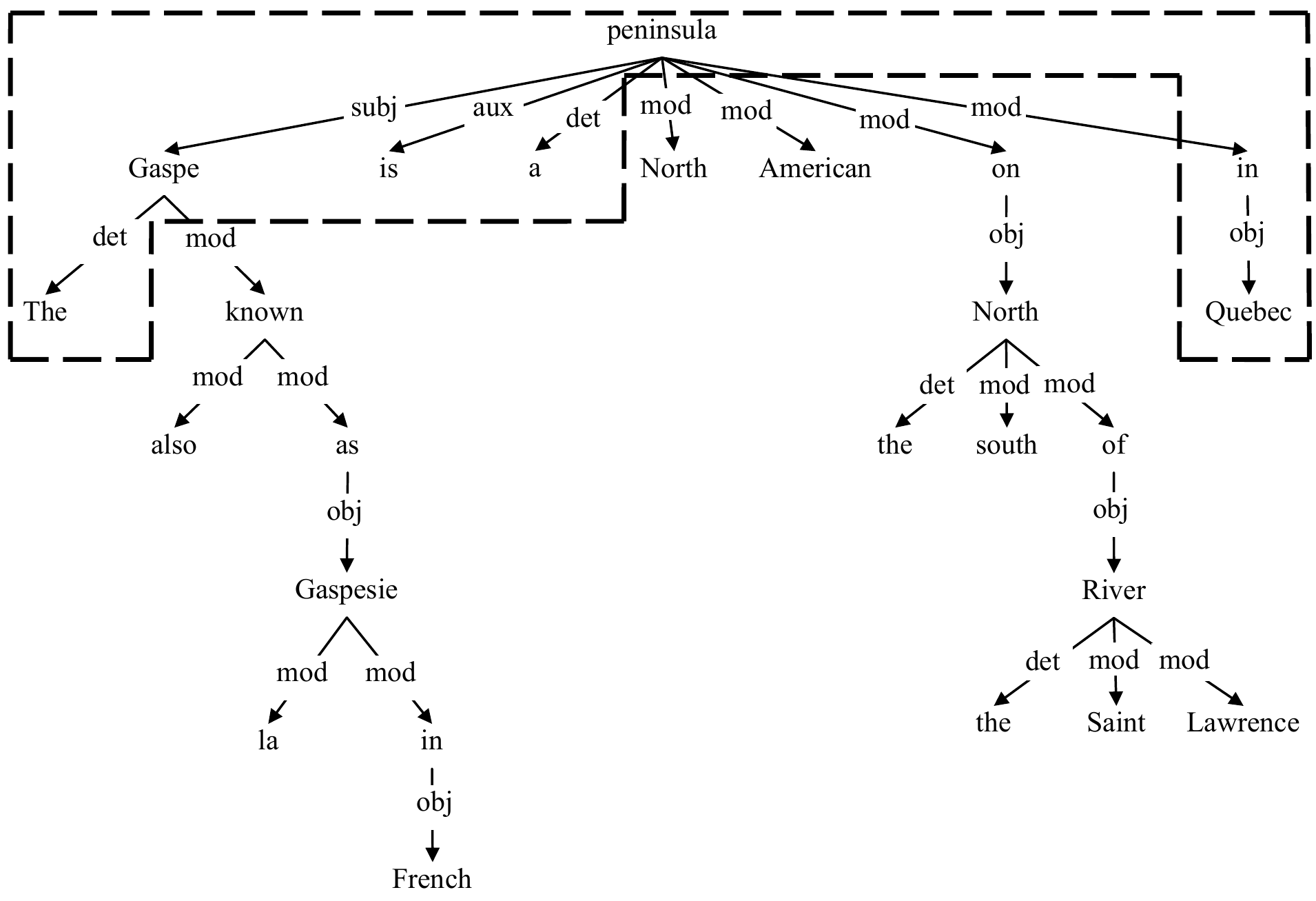}
\caption{An example of how dependency trees may make it easier to match a short sentence (subtree inside the dashed line) to a part of a longer one.}
\label{fig:dependency_graphs_partial_match}
\end{centering}
\end{figure}

In textual entailment recognition, 
one may compare $H$'s parse tree against subtrees of $T$'s parse tree  \cite{iftene-balahurdobrescu:2007:WTEP,Zanzotto2009}. 
It may be possible to match 
$H$'s tree against a 
single subtree of 
$T$, in effect a single syntactic window on $T$, as illustrated in Figure \ref{fig:dependency_graphs_partial_match}, which shows the dependency trees of \pref{quebec1}--\pref{quebec2}; 
recall that \pref{quebec2} does not match a single window of \pref{quebec1} at the surface string level.\footnote{
Figure \ref{fig:dependency_graphs_partial_match} is based on the output of Stanford's parser. One might argue that ``North'' should modify ``American''.}
 This is also a further example of how operating at a higher level than 
surface strings may reveal similarities that may be less clear at lower levels. Another example is \pref{israel1}--\pref{israel2}; although \pref{israel1} includes verbatim \pref{israel2}, it does not textually entail \pref{israel2}.\footnote{Modified example from Haghighi et al.'s \citeyear{Haghighi2005} work.} This is clear when one compares the syntactic representations of the two sentences: Israel is the subject of ``was established'' in \pref{israel2}, but not in \pref{israel1}. The difference, however, is not evident at the surface string level, and a sliding window of \pref{israel1} would match exactly \pref{israel2}, wrongly suggesting a textual entailment.  

\begin{examps}
\item $T$: The National Institute for Psychobiology in Israel was established in 1979.\label{israel1}
\item $H$: Israel was established in 1979.\label{israel2}
\end{examps}

Similar arguments can be made in favour of computing similarities at the semantic level \cite{Qiu2006}; for example, both the active and passive forms of a sentence may be mapped to the same logical formula, making their similarity clearer than at the surface or syntax level. The syntactic or semantic representations of the input expressions, however, cannot always be computed accurately (e.g., due to parser errors), which may introduce noise; and, possibly because of the noise, methods that operate at the 
syntactic or semantic level do not necessarily outperform in practice methods that operate on surface strings \cite{Wan2006,burchardt-EtAl:2007:WTEP,Burchardt2009}.

\subsection{Recognition via Similarity Measures Operating on  Symbolic Meaning Representations}

Paraphrases may also be recognized by computing similarity measures on graphs whose edges do not correspond to syntactic dependencies, but 
reflect semantic relations mentioned in the input expressions \cite{Haghighi2005}, for example the relation between a buyer and the entity bought. Relations of this kind may be identified by applying semantic role labeling methods 
\cite{Marquez2008} to the input language expressions. It is also possible to compute similarities between meaning representations that are based on 
FrameNet's frames \cite{burchardt-EtAl:2007:WTEP}. 
The latter approach has the advantage that semantically related expressions may invoke the same frame (as with ``announcement'', ``announce'', ``acknowledge'') or interconnected frames (e.g., FrameNet links the frame invoked by ``arrest'' to the frame invoked by ``trial'' via a path of temporal precedence relations), making similarities and implications easier to capture \cite{Burchardt2009}.\footnote{Consult, for example, the work of Erk and Pad\'{o} \citeyear{Erk2006} for a description of a system that can annotate texts with FrameNet frames. 
The \textsc{fate} corpus \cite{Burchardt2008}, a version of the \rte2 test set \cite{Bar-Haim2006} with FrameNet annotations, is publicly available.}
The prototypical semantic roles that PropBank \cite{Palmer2005} associates with each verb may also be used in a similar manner, instead of FrameNet's frames. 
Similarly, in the case of textual entailment recognition, one may compare $H$'s semantic representation (e.g., semantic graph or frame) to parts of $T$'s representation.

WordNet \cite{fellbaum98}, automatically constructed collections of near synonyms \cite{Lin1998b,Moore2001,Brocket2005}, or resources like \textsc{nomlex} \cite{Meyers1998} and \textsc{CatVar} \cite{Habash2003} that provide nominalizations of verbs and other derivationally related words across different \pos categories (e.g., ``to invent'' and ``invention''), can be used to match synonyms, hypernyms--hyponyms, or, more generally, semantically related words across the two input expressions. According to WordNet, in \pref{gbt1}--\pref{gbt2} ``shares'' is a direct hyponym (more specific meaning) of ``stock'', ``slumped'' is a direct hyponym of ``dropped'', and ``company'' is an indirect hyponym (two levels down) of ``organization''.\footnote{Modified example from the work of Tsatsaronis \citeyear{Tsatsaronis2009}} By treating semantically similar words (e.g., synonyms, or hypernyms-hyponyms up to a small hierarchical distance) as identical \cite{rinaldi03,Finch2005,Tatu2006,iftene-balahurdobrescu:2007:WTEP,Malakasiotis2009,Harmeling2009}, or by considering (e.g., counting) semantically similar words across the two input language expressions \cite{Brocket2005,Bos2005}, paraphrase recognizers may be able to cope with paraphrases that have very similar meanings, but very few or no common words.

\begin{examps}
\item The shares of the company dropped.\label{gbt1}
\item The organization's stock slumped.\label{gbt2}
\end{examps}

\noindent In textual entailment recognition, it may be desirable to allow the words of $T$ to be more distant hyponyms of the words of $H$, compared to paraphrase recognition. For example, ``$X$ is a computer'' textually entails ``$X$ is an artifact'', and ``computer'' is a hyponym of ``artifact'' four levels down.

Measures that exploit WordNet (or similar resources) and compute the semantic similarity between two words or, more generally, two texts have also been proposed 
\cite{Leacock:98,Lin:98,Resnik:99,Budanitsky:06,Tsatsaronis2010}.\footnote{Pedersen's WordNet::Similarity package implements many of these measures.} Some of them are directional, making them more suitable to textual entailment recognition \cite{Mihalcea2005b}. 
Roughly speaking, measures of this kind consider (e.g., sum the lengths of) the paths in WordNet's hierarchies (or similar resources) that connect the senses of corresponding (e.g., most similar) words across the two texts. 
They may also take  into account information such as the frequencies of the words in the two texts and how rarely they are encountered in documents of a large collection (inverse document frequency). The rationale is that frequent words of the input texts that are rarely used in a general corpus are more important, as in information retrieval; hence, the paths that connect them should be assigned greater weights. Since they often consider paths between word \emph{senses}, many of these measures would ideally be combined with word sense disambiguation \cite{Yarowsky2000,Stevenson2003,Kohomban:05,Navigli:08},
which is not, however, always accurate enough for practical purposes.

\subsection{Recognition Approaches that Employ Machine Learning}

\begin{figure}
\begin{centering}
\includegraphics[width=0.7\textwidth]{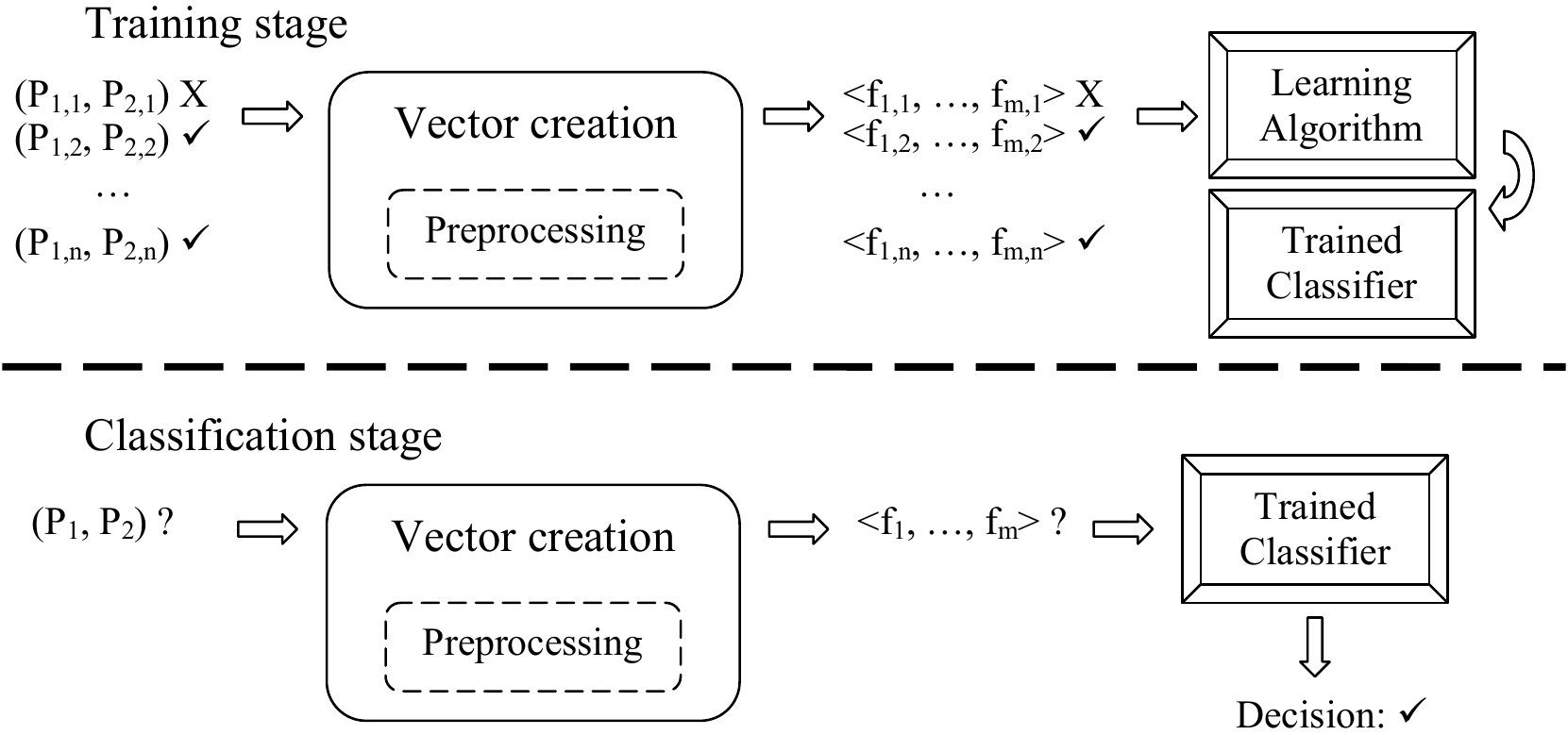}
\caption{Paraphrase and textual entailment recognition via supervised machine learning.}
\label{fig:paraphrase_recognition}
\end{centering}
\end{figure}

Multiple similarity measures, possibly computed at different levels (surface strings, syntactic or semantic representations) may be combined by using machine learning \cite{Mitchell1997,Alpaydin2004},
as illustrated in Figure \ref{fig:paraphrase_recognition}.\footnote{\textsc{weka} \cite{Witten2005} provides implementations of several well known machine learning algorithms, including \textsc{c4.5} \cite{Quinlan1993}, Naive Bayes \cite{Mitchell1997}, \svms \cite{Vapnik1998,Cristianini2000,Joachims2002}, and AdaBoost 
\cite{Freund1995,Friedman2000}. More efficient implementations of \svms, such as \textsc{libsvm} and \textsc{svm-light}, are also available. Maximum Entropy classifiers 
are also very effective; see chapter 6 of the book ``Speech and Language Processing'' \cite{jurafsky2008} for an introduction; Stanford's implementation is frequently used.}
Each pair of input language expressions $\left<P_1, P_2\right>$, 
i.e., each pair of expressions we wish to check if they are paraphrases or a correct textual entailment pair, is represented by a feature vector $\left<f_1,\dots,f_m\right>$. 
The vector contains the scores of 
multiple similarity measures applied to the pair, and possibly other features. For example, many systems 
also include features that check for polarity differences across the two input expressions, as in ``this is \emph{not} a confirmed case of rabies'' vs.\ ``a case of rabies \emph{was} confirmed'', or modality differences, as in ``a case \emph{may} have been confirmed'' vs.\ ``a case \emph{has} been confirmed'' \cite{Haghighi2005,iftene-balahurdobrescu:2007:WTEP,tatu-moldovan:2007:WTEP}. Bos and Markert \citeyear{Bos2005} also include features indicating if a theorem prover has managed to prove that the logical representation of one of the input expressions entails the other or contradicts it. A supervised machine learning algorithm trains a classifier on manually classified (as correct or incorrect) vectors corresponding to training input pairs. Once trained, the classifier can classify unseen pairs as correct or incorrect paraphrases or textual entailment pairs by examining their features 
\cite{Bos2005,Brocket2005,Zhang2005,Finch2005,Wan2006,burchardt-EtAl:2007:WTEP,Hickl2008,Malakasiotis2009,Nielsen2009}. 

A preprocessing stage is commonly applied to each input pair of language expressions, before converting it to a feature vector \cite{Zhang2005}. Part of the preprocessing may provide information that is required to compute the features; for example, this is when a 
\pos tagger or a parser would be applied.\footnote{Brill's  \citeyear{Brill1992} \pos tagger is well-known and publicly available. Stanford's tagger \cite{Toutanova2003} is another example of a publicly available \pos tagger. 
Commonly used parsers include Charniak's \citeyear{Charniak2000}, Collin's \citeyear{Collins2003}, the Link Grammar Parser \cite{Sleator1993}, \textsc{minipar}, a principle-based parser \cite{Berwick91} very similar to \textsc{principar} \cite{Lin1994}, MaltParser \cite{Nivre2007}, and Stanford's parser 
\cite{Klein2003}.} The preprocessing may also normalize the input pairs; for example, a stemmer may be applied; dates may be converted to a consistent format; names of persons, organizations, locations etc.\ may be tagged by their semantic categories using a named entity recognizer; pronouns or, more generally, referring expressions, may be replaced by the expressions they refer to \cite{Hobbs1986,Lappin1994,Mitkov2003,Molla2003,Yang2008}; and morphosyntactic variations may be normalized (e.g., passive sentences may be converted to active ones).\footnote{Porter's stemmer \citeyear{Porter1997} is well-known. An example of a publicly available named-entity recognizer is Stanford's.}

Instead of mapping each $\left<P_1, P_2\right>$ pair to a feature vector that contains mostly scores measuring the similarity between $P_1$ and $P_2$, it is possible to use vectors that encode directly parts of $P_1$ and $P_2$, or parts of their syntactic or semantic representations. Zanzotto et al.\ \citeyear{Zanzotto2009} project each $\left<P_1, P_2\right>$ pair to a vector that, roughly speaking, contains as features all the fragments of $P_1$ and $P_2$'s parse trees. Leaf nodes corresponding to identical or very similar words (according to a WordNet-based similarity measure) across $P_1$ and $P_2$ are replaced by co-indexed slots, to allow the features to be more general. Zanzotto et al.\ define a measure (actually, different versions of it) that, in effect, computes the similarity of two pairs $\left<P_1, P_2\right>$ and $\left<P'_1, P'_2\right>$ by counting the parse tree fragments (features) that are shared by $P_1$ and $P'_1$, and those shared by $P_2$ and $P'_2$. The measure is used as a kernel in an Support Vector Machine (\svm) that learns to separate positive textual entailment pairs $\left<P_1, P_2\right> = \left<T, H\right>$ from negative ones. A (valid) kernel can be thought of as a similarity measure that projects two objects to a highly dimensional vector space, where it computes the inner product of the projected objects; efficient kernels compute the inner product directly from the original objects, without computing their projections to the highly dimensional vector space \cite{Vapnik1998,Cristianini2000,Joachims2002}. In Zanzotto et al.'s work, each object is a $\left<T, H\right>$ pair, and its projection is the vector that contains  all the parse tree fragments of $T$ and $H$ as features. Consult, for example, the work of Zanzotto and Dell' Arciprete \citeyear{Zanzotto2009EMNLP} and  Moschitti \citeyear{Moschitti2009} for further discussion of kernels that can be used in paraphrase and textual entailment recognition.

\subsection{Recognition Approaches Based on Decoding} 
\label{recognition_sequences}

Pairs of paraphrasing or textual entailment expressions (or templates) like \pref{sweets0}, often called \emph{rules}, that may have been produced by extraction mechanisms (to be discussed in Section \ref{sec:extraction}) can be used by recognizers much as, and often in addition to synonyms and hypernyms-hyponyms. 

\begin{examps}
\item ``$X$ is fond of $Y$'' $\approx$ ``$X$ likes $Y$''\label{sweets0}
\end{examps}

\noindent Given the paraphrasing rule of \pref{sweets0} and the information that ``child'' is a synonym of ``kid'' and ``candy'' a hyponym of ``sweet'', a recognizer could figure out that \pref{sweets1} textually entails \pref{sweets4} by gradually transforming \pref{sweets1} to \pref{sweets4} as shown below.\footnote{Modified example from  Bar-Haim et al.'s \citeyear{BarHaim2009} work.}

\begin{examps}
\item Children are fond of sweets.\label{sweets1}
\item Kids are fond of sweets. 
\item Kids like sweets.
\item Kids like candies.\label{sweets4}
\end{examps}

Another recognition approach, then, is to search for a sequence of rule applications or other  transformations (e.g., replacing synonyms, or hypernyms-hyponyms) that turns one of the input expressions (or its syntactic or semantic representation) to the other. We call this search \emph{decoding}, because it is similar to the decoding stage of machine translation (to be discussed in Section \ref{sec:generation}), where a sequence of transformations that turns a source-language expression into a target-language expression is sought. In our case, if a sequence is found, the two input expressions constitute a positive paraphrasing or textual entailment pair, depending on the rules used; otherwise, the pair is negative. If each rule is associated with a confidence score (possibly learnt from a training dataset) that reflects the degree to which the rule preserves the original meaning in paraphrase recognition, or the degree to which we are confident that it produces an entailed expression, we may search for the sequence of transformations with the maximum score (or minimum cost), much as in approaches that compute the minimum (string or tree) edit distance between the two input expressions. The pair of input expressions can then be classified as positive if the maximum-score sequence exceeds a confidence threshold \cite{Harmeling2009}. One would also have to consider the contexts where rules are applied, because a rule may not be valid in all contexts, for instance because of the different possible senses of the words it involves. A possible solution is to associate each rule with a vector that represents the contexts where it can be used (e.g., a vector of frequently occurring words in training contexts where the rule applies), and use a rule only in contexts that are similar to its associated context vector; with slotted rules, one can also model the types of slot values (e.g., types of named entities) the rule can be used with the work of Pantel, Bhagat, Coppola, Chklovski, and Hovy \citeyear{Pantel2007}, and Szpektor, Dagan, Bar-Haim, and Goldberger \citeyear{Szpektor2008}.

Resouces like WordNet and extraction methods, however, provide thousands or millions of  rules, giving rise to an exponentially large number of transformation sequences to consider.\footnote{Collections of transformation rules and resources that can be used to obtain such rules are listed at the \textsc{acl} Textual Entailment Portal. Mirkin et al.\ \citeyear{Mirkin2009b} discuss how to evaluate collections of textual entailment rules.} When operating at the level of semantic representations, the sequence sought is in effect a proof that the two input expressions are paraphrases or a valid textual entailment pair, and it may be obtained by exploiting theorem provers, as discussed earlier. Bar-Haim et al.\ \citeyear{BarHaim2007} discuss how to search for sequences of transformations, seen as proofs at the syntactic level, when the input language expressions and their reformulations are represented by dependency trees. In subsequent work \cite{BarHaim2009}, they introduce compact forests, a data structure that allows the dependency trees of multiple intermediate reformulations to be represented by a single graph, to make the search more efficient. They also combine their approach with an \svm-based recognizer; sequences of transformations are used to bring $T$ closer to $H$, and the \svm recognizer is then employed to judge if the transformed $T$ and $H$ consitute a positive textual entailment pair or not.

\subsection{Evaluating Recognition Methods}
\label{recognition_evaluation}

Experimenting with paraphase 
and textual entailment recognizers requires datasets containing both positive and negative input pairs. When using discriminative classifiers (e.g., \svms), the negative training pairs must ideally be near misses, otherwise they may be of little use \cite{Schohn2000,Tong2002}. Near misses can also make the test data more challenging. 

The most widely used benchmark dataset for paraphrase recognition is the Microsoft Research (\msr) Paraphrase Corpus. It contains 5,801 pairs of sentences obtained from clusters of online news articles referring to the same events \cite{Dolan2004,Dolan2005b}. The pairs were initially filtered by heuristics, which require, for example, the word edit distance of the two sentences in each pair to be neither too small (to avoid nearly identical sentences) nor too large (to avoid too many negative pairs); and both sentences to be among the first three of articles from the same cluster (articles referring to the same event), the rationale being that initial sentences often summarize the events. The candidate paraphrase pairs were then filtered by an \svm-based paraphrase recognizer \cite{Brocket2005}, trained on separate manually classified pairs obtained in a similar manner, which was biased to overidentify paraphrases. Finally, human judges annotated the remaining sentence pairs as paraphrases or not. 
After resolving disagreements, approximately 67\% of the 5,801 pairs were judged to be paraphrases. The dataset is divided in two non-overlapping parts, for training (30\% of all pairs) and testing (70\%). 
Zhang and Patrick \citeyear{Zhang2005} and others have pointed out that the heuristics that were used to construct the corpus may have biased it towards particular types of paraphrases, excluding for example paraphrases that do not share any common words.

\begin{table}
{\footnotesize
\begin{center}
\begin{tabular}{|l|c|c|c|c|}
\hline \textit{method}    & \textit{accuracy (\%)} & \textit{precision (\%)} & \textit{recall (\%)} & $F$-\textit{measure (\%)}\\%
\hline Corley \& Mihalcea \citeyear{Mihalcea2005b}    & 71.5 & 72.3 & 92.5  & 81.2\\
\hline Das \& Smith       \citeyear{Das2009}          & 76.1 & 79.6 & 86.1  & 82.9\\
\hline Finch et al.       \citeyear{Finch2005}        & 75.0 & 76.6 & 89.8  & 82.7\\
\hline Malakasiotis       \citeyear{Malakasiotis2009} & 76.2 & 79.4 & 86.8  & 82.9\\
\hline Qiu et al.         \citeyear{Qiu2006}          & 72.0 & 72.5 & 93.4  & 81.6\\
\hline Wan et al.         \citeyear{Wan2006}          & 75.6 & 77.0 & 90.0  & 83.0\\
\hline Zhang \& Patrick   \citeyear{Zhang2005}        & 71.9 & 74.3 & 88.2  & 80.7\\
\hdashline
\basea                                                & 66.5 & 66.5 & 100.0 & 79.9\\
\hline \baseb                                         & 69.0 & 72.4 & 86.3  & 78.8\\
\hline
\end{tabular}
\smallskip
\caption{Paraphrase recognition results on the \msr corpus.} 
\label{tbl:msr_results}
\end{center}
} 
\end{table}

Table \ref{tbl:msr_results} lists all the published results of paraphrase recognition experiments on the \msr corpus we are aware of. We include two baselines we used: \basea classifies all pairs as paraphrases; \baseb classifies two sentences as paraphrases when their surface word edit distance is below a threshold, tuned on the training part of the corpus. Four commonly used evaluation measures are used: 
accuracy, precision, recall, and \fm with equal weight on precision and recall. These measures are defined below. \textit{TP} (true positives) and \textit{FP} (false positives) are the numbers of 
pairs that have been correctly or incorrectly, respectively, classified as positive (paraphrases). \textit{TN} (true negatives) and \textit{FN} (false negatives) are the
numbers of pairs that have been correctly or incorrectly, respectively, classified as negative (not paraphrases). 

\[
\begin{array}{cc}
\textit{precision}  = \frac{\textit{TP}}{\textit{TP} + \textit{FP}}, &
\textit{recall}  = \frac{\textit{TP}}{\textit{TP} + \textit{FN}}, 
\vspace{3mm}\\
\textit{accuracy}  =  
\frac{\textit{TP} + \textit{TN}}{\mathit{TP} + \mathit{TN} + \mathit{FP} + \mathit{FN}}, &
\textit{\fm}  = 
\frac{2 \cdot \textit{precision} \cdot \textit{recall}}{\textit{precision} + \textit{recall}}
\end{array}
\]

\noindent All the systems of Table \ref{tbl:msr_results} have better recall than precision, which implies they tend to over-classify pairs as paraphrases, possibly because the sentences of each pair have at least some common words and refer to the same event. Systems with higher recall tend to have lower precision, and vice versa,
as one would expect. The high \fm of \basea is largely due to its perfect recall; its precision is significantly lower, compared to the other systems. \baseb, which uses only string edit distance, is a competitive baseline for this corpus. 
Space does not permit listing published evaluation results of all the paraphrase recognition methods that we have discussed. Furthermore, comparing results obtained on different datasets is not always meaningful.

For textual entailment recognition, the most widely used benchmarks are those of the \rte challenges. As an example, the \rte-3 corpus contains 1,600 $\left<T, H\right>$ pairs (positive or negative). Four application scenarios where textual entailment recognition might be useful were considered: information extraction, information retrieval, question answering, and summarization. There are 200 training and 200 testing pairs for each scenario; Dagan et al.\ \citeyear{Dagan2009} explain how they were constructed.
The \rte-4 corpus was constructed in a similar way, but it contains only test pairs, 250 for each of the four scenarios. A further difference is that in \rte-4 the judges classified the pairs in three classes: true entailment pairs, false entailment pairs where $H$ contradicts $T$ \cite{Harabagiu2006b,deMarneffe2008}, and false pairs where reading $T$ does not lead to any conclusion about $H$; a similar pilot task was included in \rte-3 \cite{Voorhees2008}.
The pairs of the latter two classes can be merged, if only two classes (true and false)
are desirable. 
We also note that \rte-3 included a pilot task requiring systems to justify their answers. Many of the participants, however, used technical or mathematical terminology in their explanations, which was not always appreciated by the human judges; also, the entailments were often obvious to the judges, to the extent that no justification was 
considered necessary \cite{Voorhees2008}.
Table \ref{tbl:rte4_results} lists the best accuracy results of \rte-4 participants (for two classes only), along with results of the two baselines described previously; precision, recall, and \fm scores are also shown, when available.
All four measures are defined as in paraphrase recognition, but positives and negatives are now textual entailment pairs.\footnote{\emph{Average precision}, borrowed from information retrieval evaluation, has also been used in the \rte challenges.
Bergmair \citeyear{Bergmair2009}, however, argues against using it in \rte challenges and proposes alternative measures.}  
Again, space does not permit listing published evaluation results of all the textual entailment recognition methods that we have discussed, and comparing results obtained on different datasets is not always meaningful. 

It is also possible to evaluate recognition methods indirectly, by measuring their impact on the performance of larger natural language processing systems (Section \ref{applications}). For instance, one could measure the difference in the performance of a \qa system, or the degree to which the redundancy of a generated summary is reduced when using paraphrase and/or textual entailment recognizers.

\begin{table}
{\footnotesize
\begin{center}
\begin{tabular}{|l|c|c|c|c|}
\hline \textit{method} & \textit{accuracy (\%)} & \textit{precision (\%)} & \textit{recall (\%)} & $F$-\textit{measure (\%)}\\%
\hline Bensley \& Hickl   \citeyear{Bensley2008} & 74.6 & --   & --    & --\\
\hline Iftene             \citeyear{Iftene2008}  & 72.1 & 65.5 & 93.2  & 76.9\\
\hline Siblini \& Kosseim \citeyear{Siblini2008} & 68.8 & --   & --    & --\\
\hline Wang \& Neumann    \citeyear{Wang2008}    & 70.6 & --   & --    & --\\
\hdashline \basea                                & 50.0 & 50.0 & 100.0 & 66.7\\
\hline \baseb                                    & 54.9 & 53.6 & 73.6  & 62.0\\
\hline
\end{tabular}
\smallskip
\caption{Textual entailment recognition results (for two classes) on the \rte-4 corpus.} \label{tbl:rte4_results}
\end{center}
} 
\end{table}

\section{Paraphrase and Textual Entailment Generation} \label{sec:generation}

Unlike recognizers, paraphrase or textual entailment \emph{generators} are given a single language expression (or template) as input, and they are required to produce as many output language expressions (or templates) as possible, such that the output expressions are paraphrases or they constitute, along with the input, correct textual entailment pairs. Most  generators assume that the input is a single sentence (or sentence template), and we adopt this assumption in the remainder of this section.

\subsection{Generation Methods Inspired by Statistical Machine Translation}
\label{generation_smt}

Many 
generation methods borrow ideas from statistical machine translation (\smt).\footnote{
For an introduction to \smt, see chapter 25 of the book ``Speech and Language Processing'' \cite{jurafsky2008}, and chapter 13 of the book ``Foundations of Statistical Natural Language Processing'' \cite{Manning1999}. For a more extensive discussion, consult the work of Koehn \citeyear{Koehn2009}.} Let us first introduce some central ideas from \smt, for the benefit of readers 
unfamiliar with them. 
\smt methods rely 
on very large bilingual or multilingual parallel corpora, for example the 
proceedings of the European parliament, without constructing meaning representations and often, at least until recently, without even constructing syntactic representations.\footnote{See Koehn's Statistical Machine Translation site for commonly used \smt corpora and tools.} Let us assume that we wish to translate a sentence $F$, whose words are $f_1, f_2, \dots, f_{\left|F\right|}$ in that order, from a foreign  language to our native language. Let us also denote by $N$ any candidate translation, whose words are $a_1, a_2, \dots, a_{\left|N\right|}$. The best translation, denoted $N^*$, is the $N$ with the maximum probability of being a translation of $F$, i.e:

{\small
\begin{equation}
N^* = \arg\max_{N}P(N|F) = 
\arg\max_{N}\frac{P(N)P(F|N)}{P(F)} = 
\arg\max_{N}P(N)P(F|N)
\label{equ:noisy_channel}
\end{equation}
}

\noindent Since $F$ is fixed, the denominator $P(F)$ above is constant and can be ignored when searching for $N^*$. $P(N)$ is called the \emph{language model}
and $P(F|N)$ 
the \emph{translation model}.

For modeling purposes, it is common to assume that $F$ was in fact originally written in our native language and it was transmitted to us via a noisy channel, which introduced various deformations. 
The possible deformations 
may include, for example, replacing a native word with one or more foreign ones, removing or inserting words, moving words to the left or right etc. The commonly used \ibm models 1 to 5 \cite{brown1993} provide an increasingly richer inventory of word deformations;
more recent phrase-based \smt systems \cite{koehn2003} also allow directly replacing entire phrases with other phrases. 
The foreign sentence $F$ can thus be seen as the result of applying a sequence of transformations $D = \left<d_1, d_2, \dots, d_{\left|D\right|}\right>$ to $N$, 
and it is common to search for the $N^*$ that maximizes \pref{equ:noisy_channel3}; this search is called \emph{decoding}. 

{\small
\begin{equation}
N^* = 
\arg\max_{N}P(N)\max_{D}P(F,D|N)
\label{equ:noisy_channel3}
\end{equation}
}

\noindent 
An exhaustive search is 
usually intractable. Hence, heuristic search algorithms (e.g., based on beam search) are usually employed
\cite{Germann2001,Koehn2004}.\footnote{A frequently used \smt system that includes decoding facilities is Moses.}

Assuming for simplicity that the individual deformations $d_i(\cdot)$ of $D$ are mutually independent, $P(F,D|N)$ can be computed as the product of the probabilities of $D$'s individual deformations.
Given a bilingual parallel corpus with words aligned across languages, we can estimate the probabilities of all possible deformations $d_i(\cdot)$.
In practice, however, parallel corpora do not indicate word alignment. Hence, it is common to find the most probable word alignment of the corpus given initial estimates of individual deformation probabilities, then re-estimate the deformation probabilities given the resulting alignment, and iterate \cite{brown1993,Och2003}.\footnote{\textsc{giza++} is often used to train \ibm models and align words.} 

The translation model $P(F,D|N)$ estimates the probability of obtaining $F$ from $N$ via $D$; we are interested in $N$s with high probabilities of 
leading to $F$. We also want, however, $N$ to be grammatical, and we use the language model $P(N)$ to check for grammaticality. $P(N)$ is the probability of encountering $N$ in 
our native language; it is estimated from a large monolingual corpus of our language, typically 
assuming that the probability of encountering word $a_i$ depends only on the preceding $n-1$ words. For $n=3$, 
$P(N)$ becomes:

{\small
\begin{equation}
P(N) = 
P(a_1) \cdot P(a_2 | a_1) \cdot P(a_3 | a_1, a_2) \cdot P(a_4 | a_2, a_3) 
\cdots 
P(a_{\left|N\right|}|a_{\left|N\right|-2}, a_{\left|N\right|-1})
\end{equation}
}
\noindent A language model typically also includes smoothening mechanisms, to cope with $n$-grams that are very rare or not present in the monolingual corpus, which would lead to $P(N) = 0$.\footnote{See chapter 4 of the book ``Speech and Language Processing'' \cite{jurafsky2008} and chapter 6 of the book ``Foundations of Statistical Natural Language Processing'' \cite{Manning1999} for an introduction to language models. \textsc{srilm} \cite{Stolcke2002} is  a commonly used tool to create language models.} 

In principle, an \smt system could be used to generate paraphrases, if it could be trained on a sufficiently large monolingual corpus of parallel texts. Both $N$ and $F$ are now sentences of the same language, but $N$ has to be different from the given $F$, and it has to convey the same (or almost the same) information. The main problem is that there are no readily available monolingual parallel corpora of the sizes that are used in \smt, to train the language model on them. One possibility is to use multiple translations of the same source texts; for example, different English translations of novels originally written in other languages \cite{barzilay01}, or multiple English translations of Chinese news articles, as in the Multiple-Translation Chinese Corpus. Corpora of this kind, however, are still orders of magnitude smaller than those used in \smt. 

To bypass the lack of large monolingual parallel corpora, Quirk et al.\ \citeyear{Quirk2004} use clusters of news articles referring to the same event. The articles of each cluster do not always report the same information and, hence, they are not parallel texts. Since they talk about the same event, however, they often contain phrases, sentences, or even longer fragments with very similar meanings; corpora of this kind are often called \emph{comparable}. From each cluster, Quirk et al.\ select pairs of similar sentences (e.g., with small word edit distance, but not identical sentences) using methods like those employed to create the \msr corpus (Section
\ref{recognition_evaluation}).\footnote{Wubben et al.\ \citeyear{Wubben2009} discuss similar methods to pair news titles. Barzilay \& Elhadad \citeyear{Barzilay2003} and Nelken \& Shieber \citeyear{Nelken2006} discuss more general methods to align sentences of monolingual comparable corpora. Sentence alignment methods for bilingual parallel or comparable corpora are discussed, for example, by Gale and Church \citeyear{Gale1993}, Melamed \citeyear{Melamed1999}, Fung and Cheung \citeyear{Fung2004}, Munteanu and Marcu \citeyear{Munteanu2006}; see also the work of Wu \citeyear{Wu2000}. 
Sentence alignment methods for parallel corpora may perform poorly on comparable corpora \cite{Nelken2006}.} The sentence pairs are then word aligned as in machine translation, and the resulting alignments are used to create a table of phrase pairs as in phrase-based \smt systems \cite{koehn2003}. A phrase pair $\left<P_1, P_2\right>$ consists of contiguous words (taken to be a phrase, though not necessarily a syntactic constituent) $P_1$ of one sentence that are aligned to different contiguous words $P_2$ of another sentence. Quirk et al.\ provide the following examples of discovered pairs.

\begin{footnotesize}
\begin{center}
\begin{tabular}{|l|l|}
\hline 
$P_1$ & $P_2$ \\
\hline \hline
injured & wounded \\
\hline
Bush administration & White House \\
\hline
margin of error & error margin \\
\hline
\dots & \dots \\
\hline
\end{tabular}
\end{center}
\end{footnotesize}

Phrase pairs that occur frequently in the aligned sentences may be assigned higher probabilities; Quirk et al.\ use probabilities returned by \ibm model 1. Their decoder first constructs a lattice that represents all the possible paraphrases of the input sentence that can be produced by replacing phrases by their counterparts in the phrase table; i.e., the possible deformations $d_i(\cdot)$ are the phrase replacements licensed by the phrase table.\footnote{Chevelu et al.\ \citeyear{Chevelu2009} discuss how other decoders could be developed especially for paraphrase generation.} Unlike machine translation, not all of the words or phrases need to be replaced, which is why Quirk et al.\ also allow a degenerate identity deformation $d_{id}(\xi) = \xi$; assigning a high probability to the identity deformation leads to more conservative paraphrases, with fewer phrase replacements. The decoder uses the probabilities of $d_i(\cdot)$ to compute $P(F, D| N)$ in equation \pref{equ:noisy_channel3}, and the language model to compute $P(N)$. The best scored $N^*$ is returned as a paraphrase of $F$; the $n$ most highly scored $N$s could also be returned.
More generally, the table of phrase pairs may also include synonyms obtained from WordNet or similar resources, or pairs of paraphrases (or templates) discovered by paraphrase extraction methods; in effect, Quirk et al.'s construction of a monolingual phrase table is a paraphrase extraction method. 
A language model may also be applied locally to the replacement words of a deformation and their context to assess whether or not the new words fit the original context \cite{Mirkin2009}.

Zhao et al.\ \citeyear{zhao2008,Zhao2009} demonstrated that combining phrase tables derived from multiple resources improves paraphrase generation. They also proposed scoring the candidate paraphrases by using an additional, application-dependent model, called the \emph{usability model}; for example, in sentence compression (Section \ref{applications}) the usability model rewards $N$s that have fewer words than $F$. Equation \pref{equ:noisy_channel3} then becomes \pref{equ:noisy_channel4}, where $U(F,N)$ is the usability model and $\lambda_i$ are weights assigned to the three models; similar weights can be used in \pref{equ:noisy_channel3}.

{\small
\begin{equation}
N^* = 
\arg\max_{N}U(F,N)^{\lambda_1}P(N)^{\lambda_2}\max_{D}P(F,D|N)^{\lambda_3}
\label{equ:noisy_channel4}
\end{equation}
}

\noindent Zhao et al.\ actually use a log-linear formulation of \pref{equ:noisy_channel4}; and they select the weights $\lambda_i$ that maximize an objective function that rewards many and correct (as judged by human evaluators) phrasal replacements.\footnote{
In a ``reluctant paraphrasing'' setting \cite{Dras1998}, for example when revising a document to satisfy length requirements, readability measures, or other externally imposed constraints, it may be desirable to use an objective function that rewards making as \emph{few} changes as possible, provided that the constraints are satisfied.  Dras \citeyear{Dras1998} discusses a formulation of this problem in terms of integer programming.} One may replace the translation model by a paraphrase recognizer (Section \ref{sec:recognition}) that returns a confidence score; in its log-linear formulation, \pref{equ:noisy_channel4} then becomes \pref{equ:noisy_channel5}, where $R(F,N)$ is the confidence score of the recognizer. 

{\small
\begin{equation}
N^* = 
\arg\max_{N}[\lambda_1\log U(F,N) +
\lambda_2\log P(N) +
\lambda_3log R(F,N)]
\label{equ:noisy_channel5}
\end{equation}
}

Including hyponyms-hypernyms or textual entailment rules (Section
\ref{recognition_sequences}) in the phrase table would generate sentences $N$ that 
textually entail or are entailed (depending on the direction of the rules
and whether we replace hyponyms by hypernyms or the reverse) by $F$. \smt-inspired methods, however, have been used mostly in paraphrase generation, not in textual entailment generation. 

Paraphrases can also be generated by using pairs of machine translation systems to translate the input expression to a new language, often called a \emph{pivot} language, and then back to the original language. The resulting expression is often different from the input one, especially when the two translation systems employ different methods. By using different pairs of machine translation systems or different pivot languages, multiple paraphrases may be obtained. Duboue and Chu-Carroll \citeyear{Duboue2006} demonstrated the benefit of using this approach to paraphrase questions, with an additional machine learning classifier to filter the generated paraphrases;
their classifier uses features such as the cosine similarity between a candidate generated paraphrase and the original question, the lengths of the candidate paraphrase and the original question, features showing whether or not both questions are of the same type (e.g., both asking for a person name), etc. An advantage of this approach is that the machine translation systems can be treated as black boxes, and they can be trained on readily available parallel corpora of different languages. A disadvantage is that translation errors from both directions may lead to poor paraphrases. We return to pivot languages in Section \ref{sec:extraction}.

In principle, the output 
of a generator may be produced by mapping the input to a representation of its meaning, a process that usually presupposes 
parsing, and by passing on the meaning representation, or new meaning representations that are logically entailed by the original one, to a natural language generation system \cite{Reiter2000,Bateman2003} to produce paraphrases or entailed language expressions. This approach 
would be similar to using  language-independent meaning representations (an ``interlingua'') in machine translation, but here the meaning representations 
would not need to be language-independent, since only one language is involved. 
An approach similar to syntactic transfer in machine translation may also be adopted \cite{McKeown1983}. In that case, the input language expression (assumed to be a sentence) is first parsed. The resulting syntactic representation is then modified in ways that preserve, or affect only slightly, the original meaning (e.g., turning a sentence from active to passive), or in ways that produce syntactic representations of entailed language expressions (e.g., pruning certain modifiers or subordinate clauses). New language expressions are then generated from the new syntactic representations, possibly by invoking the surface realization components of a natural language generation system.
Parsing, however, the input expression may introduce errors, and producing a correct meaning representation of the input, when this is required, may be far from trivial. Furthermore, the natural language generator may be capable of producing language expressions of only a limited variety, missing possible paraphrases or entailed language expressions.
This is perhaps why meaning representation and syntactic transfer do not seem to be currently popular in paraphrase and textual entailment generation.

\subsection{Generation Methods that Use Bootstrapping}
\label{generation_bootstrapping}

When the input and output expressions are slotted templates, it is possible to apply bootstrapping to a large monolingual corpus (e.g., the entire Web), instead of using machine translation methods. Let us assume, for example, that we wish to generate paraphrases of \pref{boot1}, and that we are given a few pairs of seed values of $X$ and $Y$, as in \pref{boot2} and \pref{boot3}.

\begin{examps}
\item $X$ is the author of $Y$.\label{boot1}
\item $\left<X=\mbox{``Jack Kerouac''}, Y=\mbox{``On the Road''}\right>$\label{boot2}
\item $\left<X=\mbox{``Jules Verne''}, Y=\mbox{``The Mysterious Island''}\right>$\label{boot3}
\end{examps}

\noindent We can retrieve from the corpus sentences that contain any of the seed pairs:

\begin{examps}
\item Jack Kerouac wrote ``On the Road''.\label{boot4}
\item ``The Mysterious Island'' was written by Jules Verne.\label{boot5}
\item Jack Kerouac is most known for his novel ``On the Road''.\label{boot8}
\end{examps}

\noindent By replacing the known seeds with the corresponding slot names, we obtain new templates: 

\begin{examps}
\item $X$ wrote $Y$.\label{boot6}
\item $Y$ was written by $X$.\label{boot7}
\item $X$ is most known for his novel $Y$.\label{boot9}
\end{examps}

In our example, \pref{boot6} and \pref{boot7} are paraphrases of \pref{boot1}; however, \pref{boot9}  
textually entails \pref{boot1}, but is not a paraphrase of \pref{boot1}.
If we want to generate paraphrases, we must keep \pref{boot6} and \pref{boot7} only; if we want to generate templates that entail \pref{boot1}, we must keep \pref{boot9} too. Some of the generated candidate templates may neither be paraphrases of, nor entail (or be entailed by) the original template. A good paraphrase or textual entailment recognizer (Section \ref{sec:recognition}) or a human in the loop would be able to filter out bad candidate templates; see also Duclaye et al.'s \citeyear{Duclaye2003} work, where Expectation Maximization \cite{Mitchell1997} is used to filter the candidate templates. Simpler filtering techniques may also be used. For example, Ravichandran et al.\ \citeyear{ravichandran02,Ravichandran2003} assign to each candidate template a pseudo-precision score; roughly speaking, the score is computed as the number of retrieved sentences that match the candidate template with $X$ and $Y$ having the values of any seed pair, divided by the number of retrieved sentences that match the template when $X$ has a seed value and $Y$ any value, not necessarily the corresponding seed value. 

Having obtained new templates, we can search the corpus for new sentences that match them; for example, sentence \pref{boot10} matches the generated template \pref{boot7}. From the new sentences, more seed values can be obtained, if the slot values correspond to types of expressions (e.g., person names) that can be recognized reasonably well, for example by using a named entity recognizer or a gazetteer (e.g., a large list of book titles); from \pref{boot10} we would obtain the new seed pair \pref{boot11}. More iterations may be used to generate more templates and more seeds, until no more templates and seeds can be discovered or a maximum number of iterations is reached.  

\begin{examps}
\item Frankenstein was written by Mary Shelley.\label{boot10}
\item $\left<X=\mbox{``Mary Shelley''}, Y=\mbox{``Frankenstein''}\right>$\label{boot11}
\end{examps}

\begin{figure}
\begin{centering}
\includegraphics[width=0.8\textwidth]{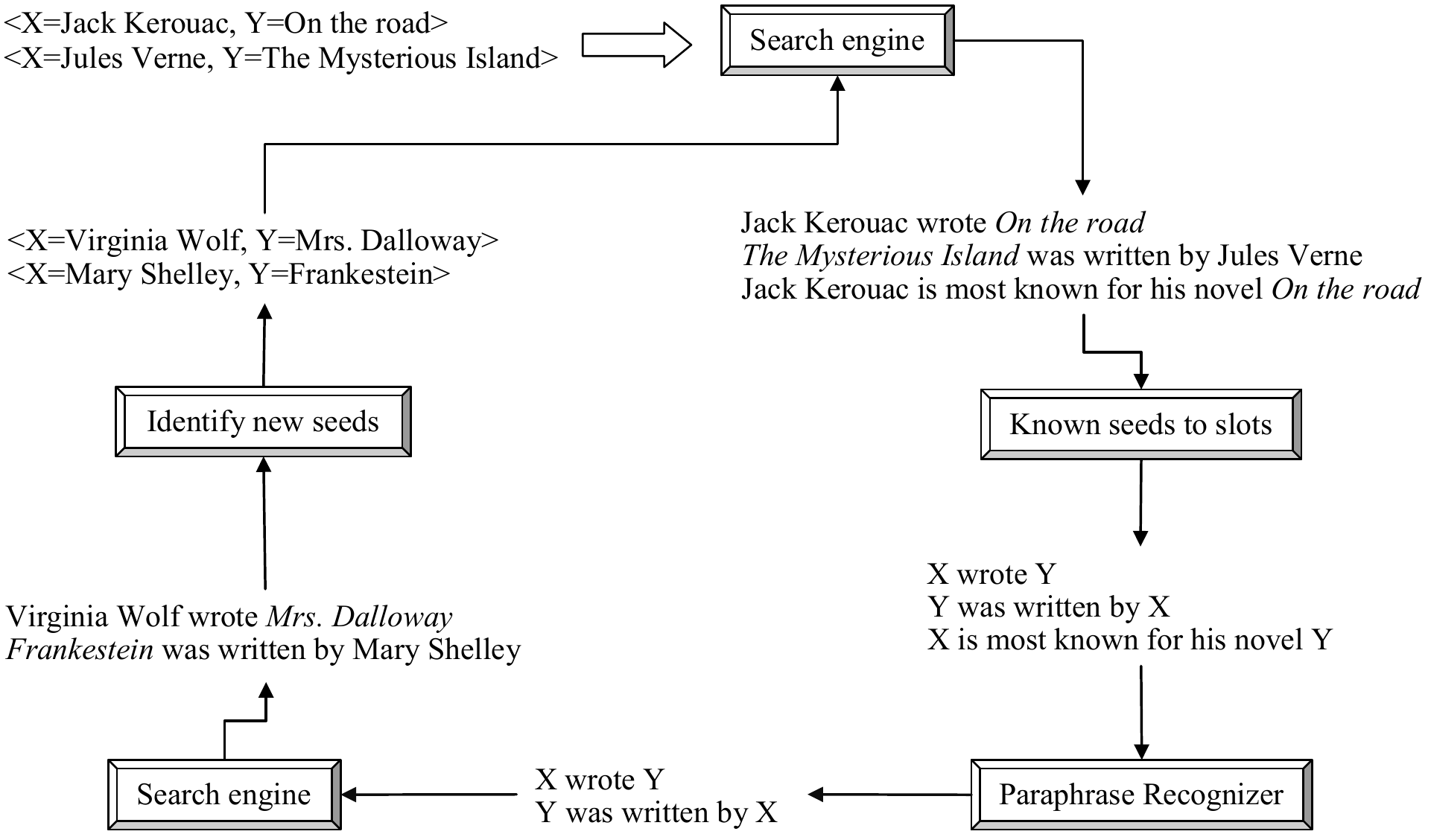}
\caption{Generating paraphrases of ``$X$ wrote $Y$'' by bootstrapping.} 
\label{fig:paraphrase_generation}
\end{centering}
\end{figure}

\noindent Figure \ref{fig:paraphrase_generation} illustrates how a bootstrapping paraphrase generator works.
Templates that textually entail or that are textually entailed by an initial template, for which seed slot values are provided, can be generated similarly, if the paraphrase recognizer is replaced by a textual entailment recognizer.

If slot values can be recognized 
reliably, we can also obtain 
the initial seed slot values automatically by retrieving directly sentences that match the original templates 
and by identifying the slot values in the retrieved sentences.\footnote{
Seed 
slot values per semantic relation can also be obtained from 
databases
\cite{Mintz2009}.}
If we are also given a mechanism to identify sentences of interest in the corpus (e.g., sentences involving particular terms, such as names of known diseases and medicines), we can also obtain the initial templates automatically, by identifying sentences of interest, identifying slot values (e.g., named entities of particular categories) in the sentences, and using the contexts of the slot values as initial templates. 
In effect, the generation task then becomes an extraction one, since we are given a corpus, but neither initial templates nor seed slot values. \tease \cite{szpektor04} is a well-known bootstrapping method of this kind, which produces textual entailment pairs, for example pairs like \pref{tease1}--\pref{tease2}, given only a monolingual (non-parallel) corpus and a dictionary of terms. \pref{tease1} textually implies \pref{tease2}, for example in contexts like those of \pref{tease3}--\pref{tease4}, but not the reverse.\footnote{Example from the work of Szpektor et al.\ \citeyear{szpektor04}.}

\begin{examps}
\item $X$ prevents $Y$ \label{tease1}
\item $X$ reduces $Y$ risk \label{tease2}
\item Aspirin prevents heart attack.\label{tease3}
\item Aspirin reduces heart attack risk.\label{tease4}
\end{examps}

\noindent \tease does not specify the directionality of the produced template pairs, for example whether \pref{tease1} textually entails \pref{tease2} or vice versa, but additional mechanisms have been proposed that attempt to guess the directionality; we discuss one such mechanism, \ledir \cite{Bhagat2007}, in Section \ref{extraction_distributional} below. Although \tease can also be used as a generator, if particular input templates are provided, we discuss it further in Section \ref{extraction_bootstrapping}, along with other bootstrapping \emph{extraction} methods, since in its full form it requires no initial templates (nor seed slot values). The reader is reminded that the boundaries between recognizers, generators, and extractors are not always clear. 

Similar bootstrapping methods have been used to generate information extraction patterns \cite{Riloff1999,Xu2007}.
Some of these methods, however, require corpora annotated with instances of particular types of events to be extracted 
\cite{Huffman1995,Riloff1996,Soderland1995,soderland1999,Muslea1999,Califf2003}, or texts that mention the target events and near-miss texts that do not \cite{Riloff1996b}. 

Marton et al.\ \citeyear{Marton2009} used a similar 
approach, but without iterations, to generate paraphrases of unknown source language phrases in a phrase-based \smt system 
(Section \ref{applications}). For each unknown phrase, they collected contexts where the phrase occurred in a monolingual corpus of the source language, and they searched for other phrases (candidate paraphrases) in the corpus that occurred in the same contexts. They subsequently produced feature vectors for both the unknown phrase and its candidate paraphrases, with each vector showing how often the corresponding phrase cooccurred with other words. The candidate paraphrases were then ranked by the similarity of their vectors to the vector of the unknown phrase. The unknown phrases were in effect replaced by their best paraphrases that the \smt system knew how to map to target language phrases, and this improved the 
\smt system's performance.  

\subsection{Evaluating Generation Methods}
\label{generation_evaluation}

In most 
generation applications, for example when rephrasing queries to a \qa 
system (Section \ref{applications}), it is desirable 
not only to produce correct 
outputs (correct paraphrases, or expressions that constitute correct textual entailment pairs along with the input), but also to produce as many correct outputs as possible.
The two goals correspond to high precision and recall, respectively.
For a particular input $s_i$, the precision $p_i$ and recall $r_i$ of a generator  can now be defined as follows (cf.\ Section \ref{recognition_evaluation}). $\textit{TP}_i$ is the number of correct outputs for input $s_i$, $\textit{FP}_i$ is the number of wrong outputs for $s_i$, and $\textit{FN}_i$ is the number of outputs for $s_i$ that have incorrectly not been generated (missed). 

\[
\begin{array}{cc}
p_i = \frac{\textit{TP}_i}{\textit{TP}_i + \textit{FP}_i}, &
r_i  = \frac{\textit{TP}_i}{\textit{TP}_i + \textit{FN}_i}
\end{array}
\]

\noindent The precision and recall scores of a method over a set of inputs $\{s_i\}$ can then be defined using micro-averaging or macro-averaging:

\[
\begin{array}{cc}
\textit{macro-precision} = \sum_{i} p_i, &
\textit{macro-recall} = \sum_{i} r_i
\end{array}
\]

\[
\begin{array}{cc}
\textit{micro-precision} = 
\frac{\sum_{i}\textit{TP}_i}
{\sum_{i}(\textit{TP}_i + \textit{FP}_i)}, &
\textit{micro-recall}  = 
\frac{\sum_{i}\textit{TP}_i}
{\sum_{i}(\textit{TP}_i + \textit{FN}_i)}
\end{array}
\]

\noindent In any case, however, recall cannot be computed in generation, because $\textit{FN}_i$ is unknown;
there are numerous correct paraphrases of an input
$s_i$ that may have been missed, and there are even more (if not infinite) language expressions that entail or are entailed by 
$s_i$.\footnote{
Accuracy (Section \ref{recognition_evaluation}) is also impossible to compute in 
this case; apart from not knowing 
$\textit{FN}_i$, the number of outputs that have correctly not been generated ($\textit{TN}_i$) is infinite.} 

Instead of reporting recall, it is common to report (along with precision) the average number of outputs, sometimes called \emph{yield}, defined below, where we assume that there are $n$ test inputs.
A better option is to report the yield at different precision levels, since there is usually a tradeoff between the two figures, 
which is controlled by parameter tuning (e.g., selecting different values of thresholds involved in the methods). 

\[
\textit{yield} = \frac{1}{n} \sum_{i=1}^{n} (\textit{TP}_i + \textit{FP}_i)
\]

Note that if we use a fixed set of test inputs $\{s_i\}$, if we store the sets $O_i^{\textit{ref}}$ of all the correct outputs that a reference generation method produces for each $s_i$, and if we treat each $O_i^{\textit{ref}}$ as the set of all \emph{possible} correct outputs that may be generated for $s_i$, then both precision and recall can be computed, and without further human effort when a new generation method, say $M$, is evaluated. $\textit{FN}_i$ is then the number of outputs in $O_i^{\textit{ref}}$ that have not been produced for $s_i$ by $M$; $\textit{FP}_i$ is the number of  $M$'s outputs for $s_i$ that are not in $O_i^{\textit{ref}}$; and $\textit{TP}_i$ is the number of $M$'s outputs for $s_i$ that are included in  $O_i^{\textit{ref}}$. Callison-Burch et al.\ \citeyear{CCB2008} propose an evaluation approach of this kind for what we call paraphrase generation. They use phrase alignment heuristics \cite{Och2003,Cohn2008} to obtain aligned phrases (e.g., ``resign'', ``tender his resignation'', ``leave office voluntarily'') from manually word-aligned sentences with the same meanings (from the Multiple-Translation Chinese Corpus). Roughly speaking, they use as $\{s_i\}$ phrases for which alignments have been found; and for each $s_i$, $O_i^{\textit{ref}}$ contains the phrases $s_i$ was aligned to. Since $O_i^{\textit{ref}}$, however, contains much fewer phrases than the possible correct paraphrases of $s_i$, the resulting precision score is a (possibly very pessimistic) lower bound, and the resulting recall scores only measure to what extent $M$ managed to discover the (relatively few) paraphrases in $O_i^{\textit{ref}}$, as pointed out by Callison-Burch et al.

To the best of our knowledge, there are no widely adopted benchmark datasets for paraphrase and textual entailment generation, unlike recognition,
and comparing results obtained on different 
datasets is not always meaningful.
The lack of generation benchmarks is probably due to the fact that although it is possible to assemble a large collection of input language 
expressions, 
it is 
practically impossible to specify in advance all the 
numerous (if not infinite) 
correct outputs 
a generator may produce,
as already discussed.
In principle, one could use a paraphrase or textual entailment recognizer to automatically judge if the output of a generator is a paraphrase of, or forms a correct entailment pair with the corresponding input expression. Current recognizers, however, are not 
yet accurate enough, and automatic evaluation measures from machine translation 
(e.g., \textsc{bleu}, Section \ref{recognition_string}) cannot be employed, exactly because their weakness is that they cannot detect paraphrases and textual entailment. An alternative, more costly solution is to use human judges, which also allows evaluating other aspects of the outputs, such as their fluency \cite{Zhao2009}, as in machine translation.
One can also evaluate the performance of a generator indirectly, by measuring its impact on the performance of larger natural language processing systems (Section \ref{applications}).

\section{Paraphrase and Textual Entailment Extraction} \label{sec:extraction}

Unlike recognition and generation methods, extraction methods are not given particular input language expressions. They typically process large corpora to extract pairs of language expressions (or templates) that constitute paraphrases or textual entailment pairs.
The generated pairs are stored to be used subsequently by recognizers and generators or other applications (e.g., as additional entries of phrase tables in \smt systems). Most extraction methods produce pairs of sentences (or sentence templates) or pairs of shorter expressions. Methods to discover synonyms, hypernym-hyponym pairs or, more generally, entailment relations between words \cite{Lin1998b,Hearst1998,Moore2001,Glickman2003,Brocket2005,Hashimoto2009,Herbelot2009} 
can be seen as performing paraphrase or textual entailment extraction restricted to pairs of single words.

\subsection{Extraction Methods Based on the Distributional Hypothesis}
\label{extraction_distributional}

A possible paraphrase extraction approach is to store all the word $n$-grams that occur in a large monolingual corpus (e.g., for $n \leq 5$), along with their left and right contexts, and consider as paraphrases 
$n$-grams that occur frequently in similar contexts. 
For example, each $n$-gram can be represented by a vector showing the words 
that typically precede or follow the $n$-gram, with the values in the vector indicating how strongly each word co-occurs with the $n$-gram; for example, pointwise mutual information values \cite{Manning1999} may be used.  
Vector similarity measures, for example cosine similarity or Lin's measure \citeyear{Lin1998b}, can then be employed to identify $n$-grams that occur in similar contexts by comparing their vectors.\footnote{
Zhitomirsky-Geffet and Dagan \citeyear{Geffet2009} discuss a bootstrapping approach, whereby the vector similarity scores (initially computed using pointwise mutual information values in the vectors) are used to improve the values in the vectors; the vector similarity scores are then re-computed.} This approach has been shown to be viable with very large monolingual corpora; Pasca and Dienes \citeyear{Pasca2005} used a Web snapshot of approximately a billion Web pages; Bhagat and Ravichandran \citeyear{Bhagat2008} used 150 \textsc{gb} of news articles and reported that results deteriorate rapidly with smaller corpora. Even if only lightweight linguistic processing (e.g., \pos tagging, without parsing) is performed, processing such large datasets requires very significant processing power, although linear computational complexity is possible with appropriate hashing of the context vectors \cite{Bhagat2008}. Paraphrasing approaches of this kind are based on Harris's Distributional Hypothesis \citeyear{harris64}, which states that words in similar contexts tend to have similar meanings. 
The bootstrapping methods of Section \ref{generation_bootstrapping} are based on a similar hypothesis that phrases (or templates) occurring in similar contexts (or with similar slot values) tend to have similar meanings, a hypothesis that can be seen as an extension of Harris's. 

\begin{figure}
\begin{centering}
\includegraphics[width=0.6\textwidth]{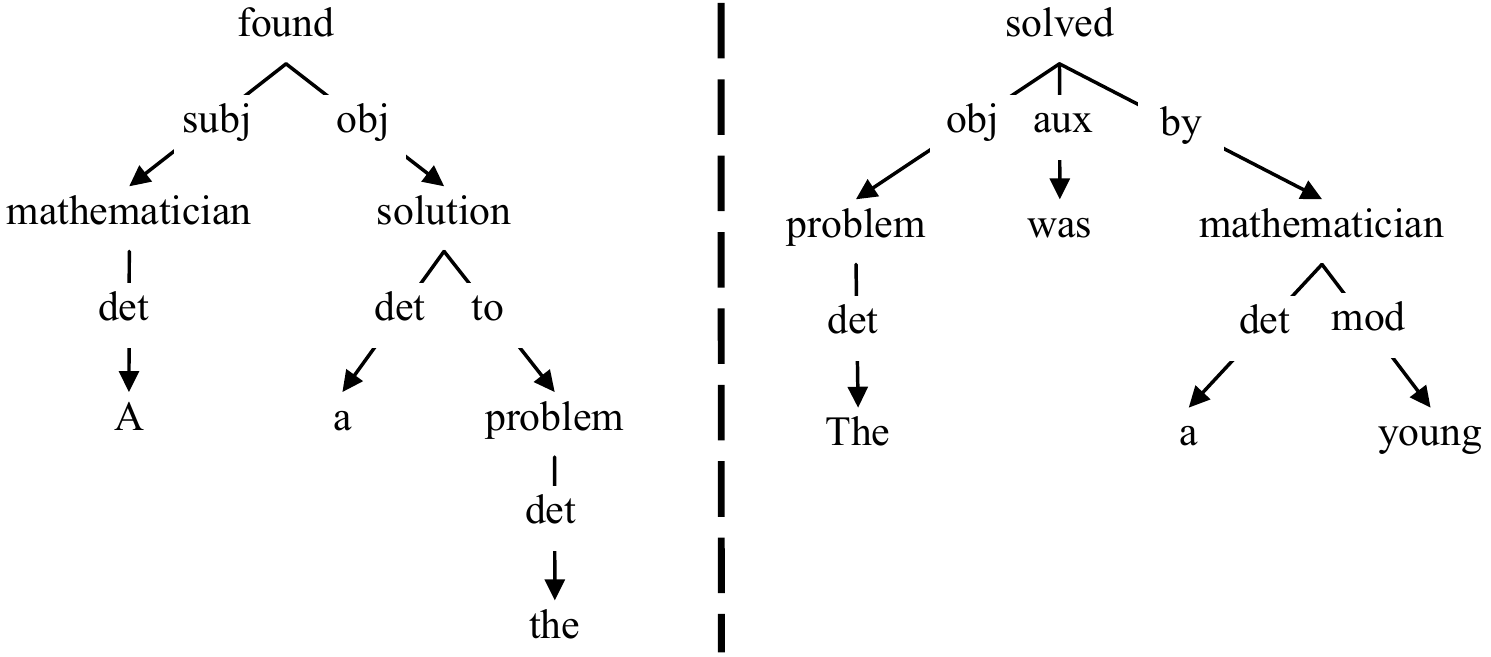}
\caption{Dependency trees of sentences \pref{lp01} and \pref{lp03}.} 
\label{fig:lin_pantel_dependency_graphs}
\end{centering}
\end{figure}

Lin and Pantel's \citeyear{lin01} well-known extraction method, called \dirt, is also based on 
the extended Distributional Hypothesis, but it operates at the syntax level. \dirt first applies a dependency grammar parser to a monolingual corpus. Parsing the corpus is generally time-consuming and, hence, smaller corpora have to be used, compared to methods that do not require parsing; Lin and Panel used 1 \textsc{gb} of news texts in their experiments. Dependency paths are then extracted from the dependency trees of the corpus. Let us consider, for example, sentences \pref{lp01} and \pref{lp03}. Their dependency trees are shown in Figure \ref{fig:lin_pantel_dependency_graphs}; the similarity between the two sentences is less obvious than in Figure \ref{fig:dependency_graphs}, because of the different verbs that are now involved. Two of the dependency paths that can be extracted from the trees of Figure \ref{fig:lin_pantel_dependency_graphs} are shown in \pref{lp02} and \pref{lp04}. The labels of the edges are augmented by the \pos-tags of the words they connect (e.g., $N$:\code{subj}:$V$ instead of simply \code{subj}).\footnote{For consistency with previous examples, we show slightly different labels than those used by Lin and Pantel.} The first and last words of the extracted paths are replaced by slots, shown as boxed and numbered \pos-tags. Roughly speaking, the paths of \pref{lp02} and \pref{lp04} correspond to the surface templates of \pref{lp02surf} and \pref{lp04surf}, respectively, but the paths are actually templates specified at the syntactic level.   

\begin{examps}
\item A mathematician found a solution to the problem.\label{lp01}
\item \fbox{$N_1$}:\code{subj}:$V$ $\leftarrow$ found $\rightarrow$ $V$:\code{obj}:$N$ $\rightarrow$ solution $\rightarrow$ $N$:\code{to}:\fbox{$N_2$} \label{lp02}
\item $N_1$ found [a] solution to $N_2$\label{lp02surf}
\item The problem was solved by a young mathematician.\label{lp03}
\item \fbox{$N_3$}:\code{obj}:$V$ $\leftarrow$ solved $\rightarrow$ $V$:\code{by}:\fbox{$N_4$} \label{lp04} 
\item $N_3$ was solved by $N_4$.\label{lp04surf} 
\end{examps}

\dirt imposes restrictions on the paths that can be extracted from the dependency trees; for example, they have to start and end with noun slots. Once the paths have been extracted, it looks for pairs of paths that occur frequently with the same slot fillers. If \pref{lp02} and \pref{lp04} occur frequently with the same  fillers (e.g., $N_1 = N_4 =$ ``mathematician'', $N_2 = N3 = $ ``problem''), they will be included as a pair in \dirt's output (with $N_1 = N_4$ and $N_2 = N3$). A measure based on mutual information \cite{Manning1999,lin01} is used to detect paths with common fillers. 

Lin and Pantel call the pairs of templates that \dirt produces ``inference rules'', but there is no directionality between the templates of each pair; the intention seems to be to produce pairs of near paraphrases. The resulting pairs are actually often textual entailment pairs, not paraphrases, and the directionality of the entailment is unspecified.\footnote{Template pairs produced by \dirt are available on-line.}
Bhagat et al.\ \citeyear{Bhagat2007} developed a method, called \ledir, to classify the template pairs $\left<P_1, P_2\right>$ that \dirt and similar methods produce into three classes: 
(i) paraphrases, (ii) $P_1$ textually entails $P_2$ and not the reverse, or (iii) $P_2$ textually entails $P_1$ and not the reverse;
with the addition of \ledir, \dirt becomes a method that extracts separately pairs of paraphrase templates and pairs of directional textual entailment templates. Roughly speaking, \ledir examines the semantic categories (e.g., person, location etc.) of the words that fill $P_1$ and $P_2$'s slots in the corpus; the categories can be obtained by following WordNet's hypernym-hyponym hierarchies from the filler words up to a certain level, or by applying clustering 
to the words of the corpus and using the clusters of the filler words as their categories.\footnote{For an introduction to clustering methods, consult chapter 14 of ``Foundations of Statistical Natural Language Processing'' \cite{Manning1999}.} If $P_1$ occurs with fillers from a substantially larger number of categories than $P_2$, then \ledir assumes $P_1$ has a more general meaning than $P_2$ and, hence, $P_2$ textually entails $P_1$; similarly for the reverse direction. If there is no substantial difference in the number of categories, $P_1$ and $P_2$ are taken to be paraphrases. 
Szpektor and Dagan \citeyear{Szpektor2008b} describe a method similar to \dirt that  produces textual entailment pairs of \emph{unary} (single slot) templates (e.g., ``$X$ takes a nap'' $\Rightarrow$ ``$X$ sleeps'') using a directional similarity measure for unary templates. 

Extraction methods based on the 
(extended) Distributional Hypothesis often produce pairs of templates that are not correct paraphrasing or textual entailment pairs, although they share many common fillers. In fact, pairs involving antonyms are frequent;  according to Lin and Pantel \citeyear{lin01}, \dirt finds ``$X$ solves $Y$'' to be very similar to ``$X$ worsens $Y$''; and the same problem has been reported in experiments with \ledir \cite{Bhagat2007} and distributional approaches that operate at the surface level \cite{Bhagat2008}.

Ibrahim et al.'s \citeyear{Ibrahim2003} method is similar to \dirt, but it assumes that a monolingual parallel corpus is available (e.g., multiple English translations of novels), whereas \dirt does not require parallel corpora. Ibrahim et al.'s method extracts pairs of dependency paths only from aligned sentences that share matching \emph{anchors}. Anchors are allowed to be only nouns or pronouns, and they match if they are identical, if they are a noun and a compatible pronoun, if they are of the same semantic category etc. In \pref{ibr01}--\pref{ibr02}, square brackets and subscripts indicate matching anchors.\footnote{Simplified example from Ibrahim et al.'s work \citeyear{Ibrahim2003}.} The pair of templates of \pref{ibr03}--\pref{ibr04} would be extracted from \pref{ibr01}--\pref{ibr02}; for simplicity, we show sentences and templates as surface strings, although the method operates on dependency trees and paths. Matching anchors become matched slots. Heuristic functions are used to score the anchor matches (e.g., identical anchors are preferred to matching nouns and pronouns) and the resulting template pairs; roughly speaking frequently rediscovered template pairs are rewarded, especially when they occur with many different anchors. 

\begin{examps}
\item The [clerk]$_1$ liked [Bovary]$_2$.\label{ibr01}
\item {[}He]$_1$ was fond of [Bovary]$_2$.\label{ibr02}
\item $X$ liked $Y$.\label{ibr03}
\item $X$ was fond of $Y$.\label{ibr04}
\end{examps}

By operating on aligned sentences of monolingual parallel corpora, Ibrahim et al.'s method may avoid, to some extent, producing pairs of unrelated templates that simply happen to share common slot fillers; 
the resulting pairs of templates are also more likely to be paraphrases, rather than simply textual entailment pairs, since they are obtained from aligned sentences of a monolingual parallel corpus. Large monolingual parallel corpora, however, are more difficult to obtain than non-parallel corpora, as already discussed. An alternative is to identify anchors in related sentences from comparable corpora 
(Section \ref{generation_smt}), which are easier to obtain. Shinyama and Sekine \citeyear{Shinyama2002} find pairs of sentences that share the same anchors within clusters of news articles 
reporting the same event. In their method, anchors are named entities (e.g., person names) identified using a named entity recognizer, or pronouns and noun phrases that refer to named entities; heuristics are employed to identify likely referents. Dependency trees are then constructed from each pair of sentences, and pairs of dependency paths are extracted from the trees by treating 
anchors as slots.

\subsection{Extraction Methods that Use Bootstrapping}
\label{extraction_bootstrapping}

Bootstrapping approaches can also be used in extraction, as in generation 
(Section \ref{generation_bootstrapping}), but with the additional complication that there is no particular input 
template
nor seed values of its slots
to start from. To address this complication, 
\tease \cite{szpektor04} starts with a lexicon of terms of a knowledge domain, for example names of diseases, symptoms etc.\ in the case of a medical domain; to some extent, such lexicons can be constructed automatically from a domain-specific corpus (e.g., medical articles) via term acquisition techniques \cite{Jacquemin2003}. 
\tease then extracts from a (non-parallel) monolingual corpus pairs of textual entailment templates that can be used with the lexicon's terms as slot fillers.
We have already shown a resulting pair of templates, \pref{tease1}--\pref{tease2}, in Section \ref{generation_bootstrapping}; we repeat it as \pref{tease1b}--\pref{tease2b} below.
Recall that \tease does not indicate the directionality of the resulting template pairs, for example whether \pref{tease1b} textually entails \pref{tease2b} or vice versa, but mechanisms like \ledir (Section \ref{extraction_distributional}) could be used to guess the directionality. 

\begin{examps}
\item $X$ prevents $Y$ \label{tease1b}
\item $X$ reduces $Y$ risk \label{tease2b}
\end{examps}

\noindent Roughly speaking, 
\tease first identifies noun phrases that cooccur frequently with each term of the lexicon, excluding very common noun phrases. It then uses the terms and their cooccurring noun phrases 
as seed slot values to obtain templates, and then the new templates to obtain more 
slot values, much as in Figure \ref{fig:paraphrase_generation}. In \tease, however, the templates are 
actually slotted dependency paths, and the method includes a stage that merges compatible templates to form more general ones.\footnote{
Template pairs produced by \tease are available on-line.} 
If particular input templates are provided, \tease can be used as a generator (Section \ref{generation_bootstrapping}).

Barzilay and McKeown \citeyear{barzilay01} also used a bootstrapping method, but to extract paraphrases from a \emph{parallel} monolingual corpus; they used multiple English translations of novels. Unlike previously discussed bootstrapping approaches, their method involves two classifiers (in effect, two sets of rules). One classifier examines the words the candidate paraphrases consist of, and a second one examines their contexts. The two classifiers use different feature sets (different views of the data), and the output of each classifier is used to improve the performance of the other one in an iterative manner; this is a case of co-training \cite{Blum1998}. More specifically, a \textsc{pos} tagger, a shallow parser, and a stemmer are first applied to the corpus, and the sentences are aligned across the different translations. Words that occur in both sentences of an aligned pair are treated as seed positive lexical examples; all the other pairs of words from the two sentences become seed negative lexical examples. From the aligned sentences \pref{BMc01}--\pref{BMc02}, we obtain three seed positive lexical examples, shown in \pref{BMc03}--\pref{BMc05}, and many more seed negative lexical examples, two of which are shown in \pref{BMc06}--\pref{BMc07}.\footnote{Simplified example from the work of Barzilay and McKeown \citeyear{barzilay01}.} Although seed positive lexical examples are pairs of identical words, as the algorithm iterates new positive lexical examples are produced, and some of them may be synonyms (e.g., ``comfort'' and ``console'') or pairs of longer paraphrases, as will be explained below.

\begin{examps}
\item He tried to comfort her.\label{BMc01}
\item He tried to console Mary.\label{BMc02}
\item $\left<\textit{expression}_1=\textrm{``he''}, \textit{expression}_2=\textrm{``he''}, +\right>$\label{BMc03}
\item $\left<\textit{expression}_1=\textrm{``tried''}, \textit{expression}_2=\textrm{``tried''}, +\right>$\label{BMc04}
\item $\left<\textit{expression}_1=\textrm{``to''}, \textit{expression}_2=\textrm{``to''}, +\right>$\label{BMc05}
\item $\left<\textit{expression}_1=\textrm{``he''}, \textit{expression}_2=\textrm{``tried''}, -\right>$\label{BMc06}
\item $\left<\textit{expression}_1=\textrm{``he''}, \textit{expression}_2=\textrm{``to''}, -\right>$\label{BMc07}
\end{examps}

The contexts of the positive (similarly, negative) lexical examples in the corresponding sentences are then used to construct positive (or negative) context rules, i.e., rules that can be used to obtain new pairs of positive (or negative) lexical examples. Barzilay and McKeown \citeyear{barzilay01} use the \pos tags of the $l$ words before and after the lexical examples as contexts, and in their experiments set $l = 3$. For simplicity, however, let us assume that $l = 2$; then, for instance, from \pref{BMc01}--\pref{BMc02} and the positive lexical example of \pref{BMc04}, we obtain the positive context rule of \pref{context1}. The rule says that if two aligned sentences contain two sequences of words, say $\lambda_1$ and $\lambda_2$, one from each sentence, and both $\lambda_1$ and $\lambda_2$ are preceded by the same pronoun, and both are followed by ``to'' and a (possibly different) verb, then $\lambda_1$ and $\lambda_2$ are positive lexical examples. Identical subscripts in the \pos tags denote identical words; for example, \pref{context1} requires both $\lambda_1$ and $\lambda_2$ to be preceded by the \emph{same} pronoun, but the verbs that follow them may be different. 

\begin{examps}
\item $\left<\textit{left}_1 = (\textit{pronoun}_1), \textit{right}_1 = (\textit{to}_1, \textit{verb}), 
\textit{left}_2 = (\textit{pronoun}_1), \textit{right}_1 = (\textit{to}_1, \textit{verb}), +\right>$
\label{context1}
\end{examps}

In each iteration, only the $k$ strongest positive and negative context rules are retained. The strength of each context rule is its precision, i.e., for positive context rules, the number of positive lexical examples whose contexts are matched by the rule divided by the number of both positive and negative lexical examples matched, and similarly for negative context rules. Barzilay and McKeown \citeyear{barzilay01} used $k = 10$, and they also discarded context rules whose strength was below 95\%. The resulting (positive and negative) context rules are then used to identify new (positive and negative) lexical examples. From the aligned \pref{BMc12}--\pref{BMc13}, the rule of \pref{context1} would figure out that ``tried'' is a synonym of ``attempted''; the two words would be treated as a new positive lexical example, shown in \pref{BMc14}. 

\begin{examps}
\item She tried to run away.\label{BMc12}
\item She attempted to escape.\label{BMc13}
\item $\left<\textit{expression}_1=\textrm{``tried''}, \textit{expression}_2=\textrm{``attempted''}, +\right>$\label{BMc14}
\end{examps}

The context rules may also produce multi-word lexical examples, like the one shown in \pref{BMc15}. The obtained lexical examples are generalized by replacing their words by their \pos tags, giving rise to paraphrasing rules. From \pref{BMc15} we obtain the positive paraphrasing rule of \pref{BMc16}; again, \pos subscripts denote identical words, whereas superscripts denote identical stems. The rule of \pref{BMc16} says that any sequence of words consisting of a verb, ``to'', and another verb is a paraphrase of any other sequence consisting of the same initial verb, ``to'', and another verb of the same stem as the second verb of the first sequence, provided that the two sequences occur in aligned sentences.

\begin{examps}
\item $\left<\textit{expression}_1=\textrm{``start to talk''}, \textit{expression}_2=\textrm{``start talking''}, +\right>$\label{BMc15}
\item $\left<\textit{generalized\_expression}_1= (\textit{verb}_0, \textit{to}, \textit{verb}^1), \textit{generalized\_expression}_2=(\textit{verb}_0, \textit{verb}^1), +\right>$\label{BMc16}
\end{examps}

The paraphrasing rules are also filtered by their strength, which is the precision with which they predict paraphrasing contexts. The remaining paraphrasing rules are used to obtain more lexical examples, which are also filtered by the precision with which they predict paraphrasing contexts. The new positive and negative lexical examples are then added to the existing ones, and they are used to obtain, score, and filter new positive and negative context rules, as well as to rescore and filter the existing ones. The resulting context rules are then employed to obtain more lexical examples, 
more paraphrasing rules, and so on, until no new positive lexical examples can be obtained from the corpus, or a maximum number of iterations is exceeded. Wang et al.\ \citeyear{Wang2009} added more scoring measures to Barzilay and McKeown's \citeyear{barzilay01} method to filter and rank the paraphrase pairs it produces, and used the extended method to extract paraphrases of technical terms from clusters of  
bug reports.

\subsection{Extraction Methods Based on Alignment}

\begin{figure}
\begin{centering}
\includegraphics[width=0.7\textwidth]{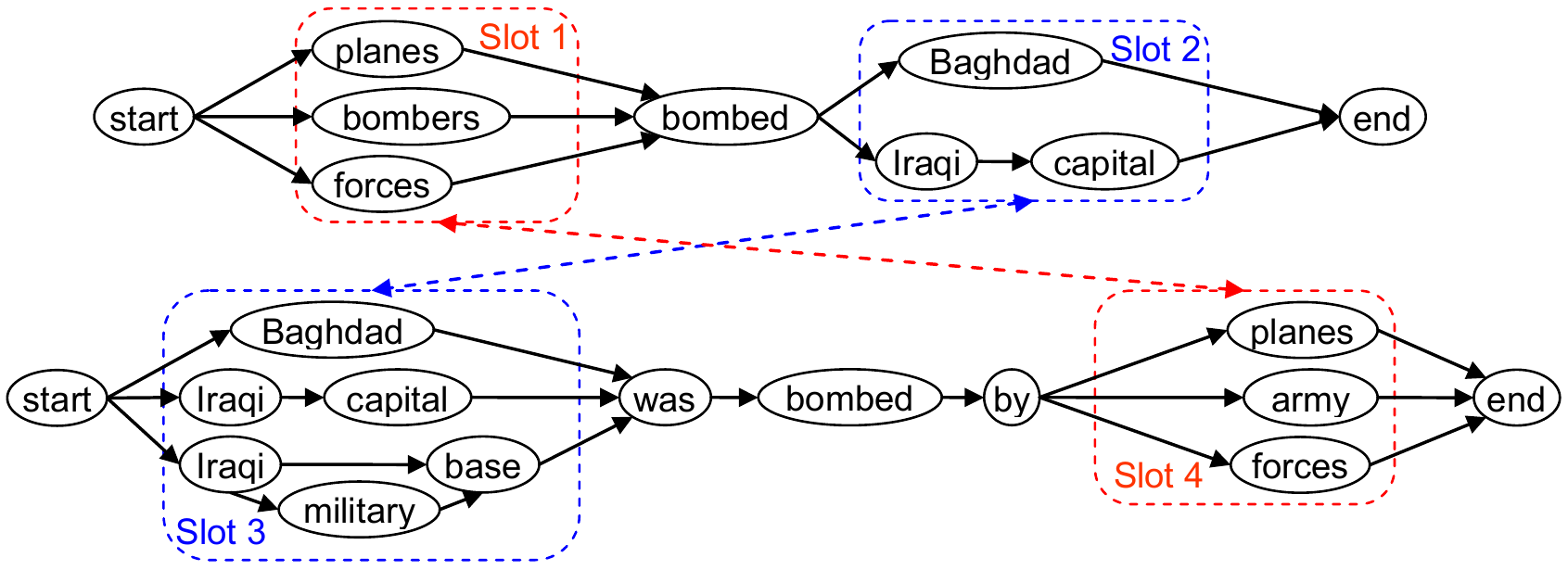}
\caption{Word lattices obtained from sentence clusters in Barzilay and Lee's method.} 
\label{fig:barzilay_lee_lattice}
\end{centering}
\end{figure}

Barzilay and Lee \citeyear{barzilay03a} used two corpora of the same genre, but from different sources (news articles from two press agencies). They call the two corpora \emph{comparable}, but they use the term with a slightly different meaning than in previously discussed methods; the sentences of each corpus were clustered separately, and each cluster was intended to contain sentences (from a single corpus) referring to events of the same \emph{type} (e.g., bomb attacks), not sentences (or documents) referring to the same events (e.g., the same particular bombing). From each cluster, 
a word lattice was produced 
by aligning the cluster's sentences with Multiple Sequence Alignment
\cite{Durbin1998,Barzilay2002EMNLP}. The solid lines of Figure \ref{fig:barzilay_lee_lattice} illustrate two possible resulting lattices, from two different clusters; we omit stop-words. Each sentence of a cluster corresponds to a path in the cluster's lattice. In each lattice, nodes that are shared by a high percentage (50\% in Barzilay and Lee's experiments) of the cluster's sentences are considered \emph{backbone nodes}. Parts of the lattice that connect otherwise consecutive backbone nodes are replaced by slots, as illustrated in Figure \ref{fig:barzilay_lee_lattice}. The two lattices of our example correspond to the surface templates \pref{msa01}--\pref{msa02}. 

\begin{examps}
\item $X$ bombed $Y$.\label{msa01}
\item $Y$ was bombed by $X$.\label{msa02}
\end{examps}

\noindent The encountered fillers of each slot are also recorded. If two slotted lattices (templates) from different corpora share many fillers, they are taken to be a pair of paraphrases (Figure \ref{fig:barzilay_lee_lattice}).
Hence, this method also uses the extended Distributional Hypothesis (Section \ref{extraction_distributional}). 

\begin{figure}
\begin{centering}
\includegraphics[width=0.5\textwidth]{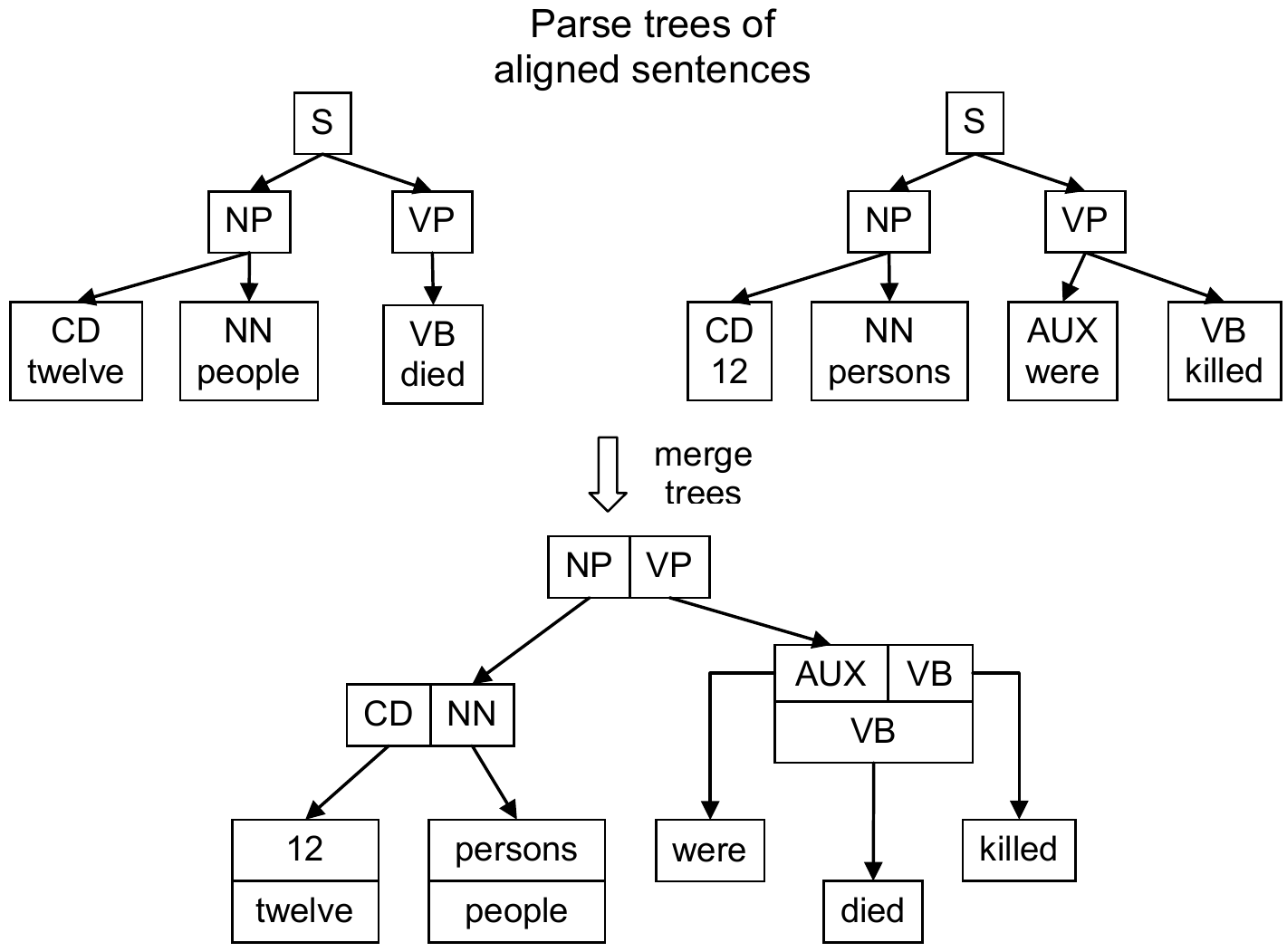}
\caption{Merging parse trees of aligned sentences in Pang et al.'s method.} 
\label{fig:pang_et_al_merge_trees}
\end{centering}
\end{figure}

Pang et al.'s method \citeyear{pang03a} produces finite state automata very similar to Barzilay and Lee's \citeyear{barzilay03a} lattices, but it requires a parallel monolingual corpus; Pang et al.\ used the Multiple-Translation Chinese Corpus (Section
\ref{generation_smt}) in their experiments. The parse trees of aligned sentences are constructed and then merged as illustrated in Figure \ref{fig:pang_et_al_merge_trees}; vertical lines inside the nodes indicate sequences of necessary constituents, whereas horizontal lines correspond to disjunctions.\footnote{Example from Pang et al.'s work \citeyear{pang03a}.} In the example of Figure \ref{fig:pang_et_al_merge_trees}, both sentences consist of a noun phrase (\texttt{NP}) followed by a verb phrase (\texttt{VP}); this is reflected to the root node of the merged tree. In both sentences, the noun phrase is a cardinal number (\texttt{CD}) followed by a noun (\texttt{NN}); however, the particular cardinal numbers and nouns are different across the two sentences, leading to leaf nodes with disjunctions. The rest of the merged tree is constructed similarly; consult Pang at al.\ for further details. Presumably one could also generalize over cardinal numbers, types of named entities etc.

\begin{figure}
\begin{centering}
\includegraphics[width=0.4\textwidth]{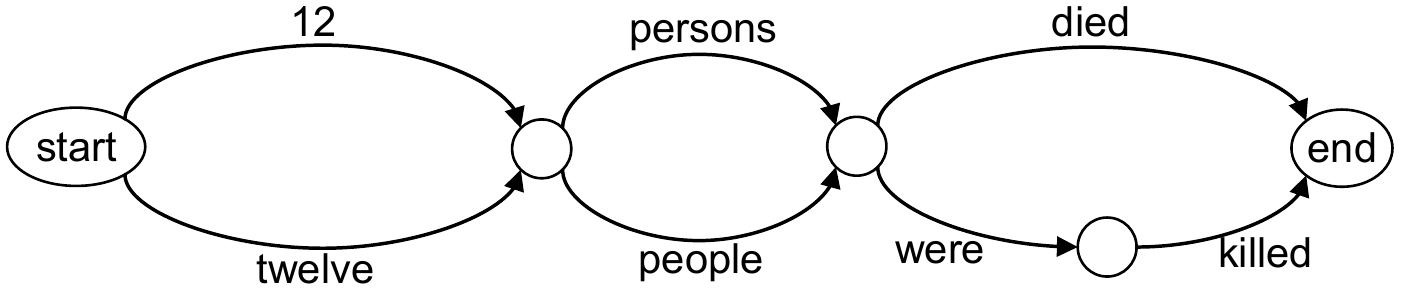}
\caption{Finite state automaton produced by Pang et al.'s method.} 
\label{fig:pang_et_al_fsa}
\end{centering}
\end{figure}

Each merged tree is then converted to a finite state automaton by traversing the tree in a depth-first manner and introducing a ramification when a node with a disjunction is encountered. Figure \ref{fig:pang_et_al_fsa} shows the automaton that corresponds to the merged tree of Figure \ref{fig:pang_et_al_merge_trees}. All the language expressions that can be produced by the automaton (all the paths from the start to the end node) are paraphrases. Hence, unlike other extraction methods, Pang et al.'s \citeyear{pang03a} method produces automata, rather than pairs of templates, but the automata can be used in a similar manner. In recognition, for example, if two strings are accepted by the same automaton, they are paraphrases; and in generation, we could look for an automaton that accepts the input expression, and then output other expressions that can be generated by the same automaton. 
As with Barzilay and Lee's \citeyear{barzilay03a} method, however, Pang et al.'s \citeyear{pang03a} method is intended to extract mostly paraphrase, not simply textual entailment pairs. 

Bannard and Callison-Burch \citeyear{Bannard2005} point out that bilingual parallel corpora are much easier to obtain, and in much larger sizes, than 
the monolingual parallel or comparable corpora that some extraction methods employ. Hence, they set out to extract paraphrases from bilingual parallel corpora commonly used in statistical machine translation (\smt). As already discussed in Section 
\ref{generation_smt}, phrase-based \smt systems employ tables whose entries show how phrases of one language may be replaced by phrases of another language; phrase tables of this kind may be produced by applying phrase alignment heuristics \cite{Och2003,Cohn2008} to word alignments produced by the commonly used \ibm models.
In the case of an English-German parallel corpus, a phrase table may contain entries like the following, which show that ``under control'' has been aligned with ``unter kontrolle'' in the corpus, but ``unter kontrolle'' has also been aligned with ``in check''; hence, ``under control'' and ``in check'' are a candidate paraphrase pair.\footnote{Example from the work of Bannard and Callison-Burch \citeyear{Bannard2005}.}

\begin{footnotesize}
\begin{center}
\begin{tabular}{|l|l|}
\hline 
English phrase & German phrase \\
\hline \hline
\dots & \dots \\
\hline 
under control & unter kontrolle \\
\hline
\dots & \dots \\
\hline
in check & unter kontrolle \\
\hline
\dots & \dots \\
\hline
\end{tabular}
\end{center}
\end{footnotesize}

More precisely, to paraphrase English phrases, Bannard and Callison-Burch \citeyear{Bannard2005} employ a \emph{pivot language} (German, in the example above) and a bilingual parallel corpus for English and the pivot language. They construct a phrase table from the parallel corpus, and from the table they estimate the probabilities $P(e|f)$ and $P(f|e)$, where $e$ and $f$ range over all of the English and pivot language phrases of the table. For example, $P(e|f)$ may be estimated as the number of entries (rows) that contain both $e$ and $f$, divided by the number of entries that contain $f$, if there are multiple rows for multiple alignments of $e$ and $f$ in the corpus, and similarly for $P(f|e)$. The best paraphrase $e_2^*$ of each English phrase $e_1$ in the table is then computed by equation \pref{pivoteq1}, where $f$ ranges over all the pivot language phrases of the phrase table $T$. 

{\small
\begin{equation}
e_2^* = 
\arg\max_{e_2 \neq e_1} P(e_2|e_1) =
\arg\max_{e_2 \neq e_1} \sum_{f \in T} P(f|e_1) P(e_2|f, e_1) \approx 
\arg\max_{e_2 \neq e_1} \sum_{f \in T} P(f|e_1) P(e_2|f) 
\label{pivoteq1}
\end{equation}
}

\noindent Multiple bilingual corpora, for different pivot languages, can be used; \pref{pivoteq1} becomes \pref{pivoteq2}, where $C$ ranges over the corpora, and $f$ now ranges over the pivot language phrases of $C$'s phrase table. 

{\small
\begin{equation}
e_2^* = 
\arg\max_{e_2 \neq e_1} \sum_C \sum_{f \in T(C)} P(f|e_1) P(e_2|f) 
\label{pivoteq2}
\end{equation}
}

Bannard and Callison-Burch \citeyear{Bannard2005}
also considered adding a language model (Section 
\ref{generation_smt}) to their method to favour paraphrase pairs that can be used interchangeably in sentences; roughly speaking, the language model assesses how well one element of a pair can replace the other in sentences where the latter occurs, by scoring the grammaticality of the sentences after the replacement. In subsequent work, Callison-Burch \citeyear{callisonburch:2008:EMNLP} extended their method to require paraphrases to have the same syntactic types, since replacing a phrase with one of a different syntactic type generally leads to an ungrammatical sentence.\footnote{An implementation of Callison-Burch's \citeyear{callisonburch:2008:EMNLP} method and paraphrase rules it produced are available on-line.} 
Zhou et al.\ \citeyear{Zhou2006b} employed a method very similar to Bannard and Callison-Burch's to extract paraphrase pairs from a corpus, and used the resulting pairs in \smt evaluation, when comparing machine-generated translations against human-authored ones.
Riezler et al.\ \citeyear{Riezler2007} adopted a similar pivot approach to obtain paraphrase pairs from bilingual phrase tables, and used the resulting pairs as paraphrasing rules to obtain paraphrases of (longer) questions submitted to a \qa system; they also used a log-linear model (Section \ref{generation_smt}) to rank the resulting question paraphrases by combining the probabilities of the invoked paraphrasing rules, a language model score of the resulting question paraphrase, and other features.\footnote{Riezler et al.\ \citeyear{Riezler2007} also employ a paraphrasing method based on an \smt system trained on question-answer pairs.}

\begin{figure}
\begin{centering}
\includegraphics[width=0.5\textwidth]{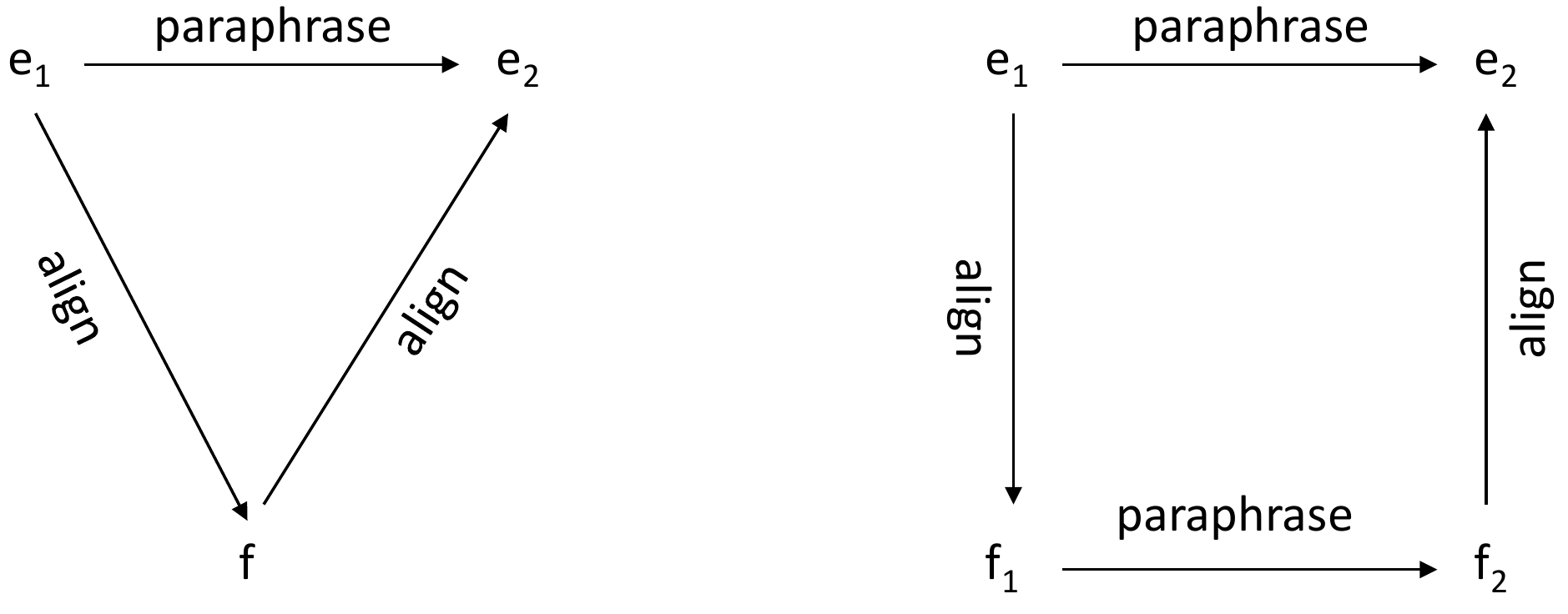}
\caption{Illustration of Zhao et al.'s pivot approaches to paraphrase extraction.}
\label{fig:pivot_paths}
\end{centering}
\end{figure}

The pivot language approaches discussed above have been shown to produce millions of paraphrase pairs from large bilingual parallel corpora. The paraphrases, however, are typically short (e.g., up to four or five words), since longer phrases are rare in phrase tables. The methods can also be significantly affected by errors in automatic word and phrase alignment \cite{Bannard2005}. To take into consideration word alignment errors, Zhao et al.\ \citeyear{zhao2008} use a log-linear classifier to score candidate paraphrase pairs that share a common pivot phrase, instead of using equations \pref{pivoteq1} and \pref{pivoteq2}. In effect, the classifier uses the probabilities $P(f|e_1)$ and $P(e_2|f)$ of  \pref{pivoteq1}--\pref{pivoteq2} as features, but it also uses additional features that assess the quality of the word alignment between $e_1$ and $f$, as well as between $f$ and $e_2$. In subsequent work, Zhao et al.\ \citeyear{Zhao2009} also consider the English phrases $e_1$ and $e_2$ to be paraphrases, when they are aligned to \emph{different} pivot phrases $f_1$ and $f_2$, provided that $f_1$ and $f_2$ are themselves a paraphrase pair in the pivot language. Figure \ref{fig:pivot_paths} illustrates the original and extended pivot approaches of Zhao et al. The paraphrase pairs $\left<f_1, f_2\right>$ of the pivot language are extracted and scored from a bilingual parallel corpus as in the original approach, by reversing the roles of the two languages. The scores of the $\left<f_1, f_2\right>$ pairs, which roughly speaking correspond to $P(f_2|f_1)$, are included as additional features in the classifier that scores the resulting English paraphrases, along with scores corresponding to $P(f_1|e_1)$, $P(e_2|f_2)$, and features that assess the word alignments of the phrases involved. 

Zhao et al.'s \citeyear{zhao2008,Zhao2009} method also extends Bannard and Callison-Burch's \citeyear{Bannard2005} by producing pairs of \emph{slotted} templates, whose slots can be filled in by words of particular parts of speech (e.g., ``\textit{Noun}$_1$ is considered by \textit{Noun}$_2$'' $\approx$ ``\textit{Noun}$_2$ considers \textit{Noun}$_1$'').\footnote{
A collection of template pairs produced by Zhao et al.'s method is available on-line.}
Hence, Zhao et al.'s patterns are more general, but a reliable parser of the language we paraphrase in is required; let us assume again that we paraphrase in English. Roughly speaking, the slots are formed by removing subtrees from the dependency trees of the English sentences and replacing the removed subtrees by the \pos tags of their roots; words of the pivot language sentences that are aligned to removed words of the corresponding English sentences are also replaced by slots. A language model is also used, when paraphrases are replaced in longer sentences. Zhao et al.'s experiments show that their method outperforms \dirt, and that it is able to output as many paraphrase pairs as the method of Bannard and Callison-Burch, but with better precision, i.e., fewer wrongly produced pairs. Most of the generated paraphrases (93\%), however, contain only one slot, and the method is still very sensitive to word alignment errors \cite{Zhao2009}, although the features that check the word alignment quality alleviate the problem. 

Madnani et al.\ \citeyear{Madnani2007} used a pivot approach similar to Bannard and Callison-Burch's \citeyear{Bannard2005} to obtain synchronous (normally bilingual) English-to-English context-free grammar rules from bilingual parallel corpora. Parsing an English text with the English-to-English synchronous rules automatically paraphrases it; 
hence the resulting synchronous rules can be used in paraphrase generation (Section \ref{sec:generation}). The rules have associated probabilities, which are estimated from the bilingual corpora. A log-linear combination of the probabilities and other features of the invoked rules is used to guide parsing. Madnani et al.\ employed the English-to-English rules to parse and, thus, paraphrase human-authored English reference translations of Chinese texts. They showed that using the additional automatically generated reference translations when tuning a Chinese-to-English \smt system improves its performance, compared to using only the human-authored references.

We note that the alignment-based methods of this section appear to have been used to extract only paraphrase pairs, not (unidirectional) textual entailment pairs.

\subsection{Evaluating Extraction Methods}

When evaluating extraction methods, we would ideally measure both their precision (what percentage of the extracted pairs are 
correct paraphrase or textual entailment pairs) and their recall (what percentage of 
all the correct pairs that could have been extracted have actually been extracted).
As in generation, however, recall cannot be computed, because the number of all correct pairs that 
could have been extracted from a large corpus (by an ideal method) is unknown. Instead, one may 
again count the number of extracted pairs
(the 
total yield of the method), possibly at different precision levels. Different extraction methods, however, produce pairs of different kinds (e.g., surface strings, slotted surface templates, or slotted dependency paths) from different kinds of corpora (e.g., monolingual or multilingual parallel or comparable corpora); hence, direct comparisons of extraction methods may be impossible. 
Furthermore, different 
scores are obtained, depending on whether the extracted pairs are considered in particular contexts or not, and whether they are required to be interchangeable in grammatical sentences \cite{Bannard2005,barzilay03a,callisonburch:2008:EMNLP,zhao2008}.
The output of an extraction method may also include pairs with relatively minor variations (e.g., active vs.\ passive, verbs vs.\ nominalizations, or variants such as ``the $X$ company bought $Y$ vs.\ ``$X$ bought $Y$''), which may cause methods that produce large numbers of minor variants to appear better than they really are;
these points also apply to the evaluation of generation methods (Section \ref{generation_evaluation}), though they have been discussed mostly in the extraction literature. Detecting and grouping such variants (e.g., turning all passives and nominalizations to active forms) may help avoid this bias and may also improve the quality of the extracted pairs by making the occurrences of the (grouped) expressions less sparse \cite{Szpektor2007}.

As in generation, in principle one could use a paraphrase or textual entailment recognizer to 
automatically score the extracted pairs.
However, recognizers 
are not 
yet accurate enough; hence, human judges are usually employed.
When extracting slotted textual entailment rules (e.g., 
``$X$ painted $Y$'' textually entails ``$Y$ is the work of $X$''), Szpektor et al.\ \citeyear{Szpektor2007b} report that human judges find it easier to agree whether or not particular instantiations of the rules (in particular contexts) are correct or incorrect, as opposed to asking them to assess directly the correctness of the rules. A better evaluation strategy, then, is to show the judges multiple sentences that match the left-hand side of each rule, along with the corresponding transformed sentences that are produced by applying the rule, and measure the percentage of these sentence pairs the judges consider correct textual entailment pairs;
this measure can be thought of as the precision of 
each individual rule.
Rules whose precision exceeds a (high) threshold can be considered correct \cite{Szpektor2007b}. 

Again, one may also evaluate extraction methods indirectly, for example by measuring how much the extracted pairs help in information extraction 
\cite{Bhagat2008,Szpektor2007,Szpektor2008b} or when expanding queries \cite{Pasca2005}, 
by measuring how well the extracted pairs, seen as paraphrasing rules, perform in phrase alignment in monolingual parallel corpora \cite{CCB2008}, or by measuring to what extent \smt or summarization evaluation measures can be improved by taking into consideration the extracted pairs \cite{CallisonBurch2006,Kauchak2006,Zhou2006b}.

\section{Conclusions} \label{section:conclusions}

Paraphrasing and textual entailment 
is currently a popular research topic. Paraphrasing can be seen as bidirectional textual entailment and, hence, similar methods are often used for both. Although both kinds of methods can be described in terms of logical entailment, they are usually intended to capture human intuitions that may not be as strict as logical entailment; and although logic-based methods have been developed, most methods operate at the surface, syntactic, or shallow semantic level, with dependency trees being a particularly popular representation. 

Recognition methods, which classify input pairs of natural language expressions 
(or templates) as correct or incorrect paraphrases or textual entailment pairs,  
often rely on supervised machine learning to combine similarity measures possibly operating at different representation levels (surface, syntactic, semantic). More recently, approaches that search for sequences of transformations that connect the two input expressions are also gaining popularity, and they exploit paraphrasing or textual entailment rules extracted from large corpora. The \rte challenges provide a 
significant thrust to recognition work, and they have helped establish benchmarks and attract more researchers.

\begin{table}
\begin{center}
{\footnotesize
\begin{tabular}{|l|c|c|c|c|c|c|}
\hline
Main ideas discussed                                    & R-TE & R-P & G-TE & G-P & E-TE & E-P \\
\hline
Logic-based inferencing                                 &  X   &  X  &      &     &      &     \\
Vector space semantic models                            &      &  X  &      &     &      &     \\
Surface string similarity measures                      &  X   &  X  &      &     &      &     \\
Syntactic similarity measures                           &  X   &  X  &      &     &      &     \\
Similarity measures on symbolic meaning representations &  X   &  X  &      &     &      &     \\
Machine learning algorithms                             &  X   &  X  &      &  X  &      &  X  \\
Decoding (transformation sequences)                     &  X   &  X  &      &  X  &      &     \\
Word/sentence alignment                                 &  X   &     &      &  X  &      &  X  \\
Pivot language(s)                                       &      &     &      &  X  &      &  X  \\
Bootstrapping                                           &      &     &  X   &  X  &  X   &  X  \\
Distributional hypothesis                               &      &     &  X   &  X  &  X   &  X  \\
Synchronous grammar rules                               &      &     &      &  X  &      &  X  \\
\hline
\end{tabular}
} 
\end{center}
\caption{Main ideas discussed and tasks they have mostly been used in.
\textsc{r}: recognition; \textsc{g}: generation, \textsc{e}: extraction; \textsc{te}: textual entailment, \textsc{p}: paraphrasing.}
\label{ideas_tasks}
\end{table}

Generation methods, meaning methods that generate paraphrases of an input natural language expression
(or template), or expressions that entail or are entailed by the input expression, are currently based mostly on bootstrapping or ideas from statistical machine translation. There are fewer publications on generation, compared to recognition (and extraction), and 
most of them focus on paraphrasing; furthermore, there are no established challenges or benchmarks,
unlike recognition. Nevertheless, generation may provide opportunities for novel research, especially to researchers with experience in statistical machine translation, who may for example wish to develop alignment or decoding techniques especially for paraphrasing or textual entailment generation.

Extraction methods extract paraphrases or textual entailment pairs 
(also called ``rules'') from corpora, usually off-line. They can be used to construct resources (e.g., phrase tables or collections of rules) that can be exploited by recognition or generation methods, or in other tasks (e.g., statistical machine translation, information extraction). Many extraction methods are based on the Distributional Hypothesis, though they often operate at different representation levels. Alignment techniques originating from statistical machine translation are recently also popular and they allow existing large bilingual parallel corpora to be exploited. Extraction methods also differ depending on whether they require parallel, comparable, or simply large corpora, monolingual or bilingual. 
As in generation, most extraction research has focused on paraphrasing, and there are no established challenges or benchmarks. 

Table \ref{ideas_tasks} summarizes the main ideas we have discussed per task, and Table \ref{ideas_resources} lists the corresponding main resources that are typically required. The underlying ideas of generation and extraction methods are in effect the same, as shown in Table \ref{ideas_tasks}, even if the methods perform different tasks; recognition work has relied on rather different ideas. Generation and extraction have mostly focused on paraphrasing, as already noted, which is why fewer ideas have been explored in generation and extraction for (unidirectional) textual entailment. 

We expect to see more interplay among recognition, generation, and extraction methods in the near future. For example, recognizers and generators may use extracted rules to a larger extent; recognizers may be used to filter candidate paraphrases or textual entailment pairs in extraction or generation approaches; and generators may help produce more monolingual parallel corpora or recognition benchmarks. We also expect to see paraphrasing and textual entailment methods being used more often in larger natural language processing tasks, including question answering, information extraction, text summarization, natural language generation, and machine translation.

\begin{table}
\begin{center}
{\footnotesize
\begin{tabular}{|l|l|}
\hline
Main ideas discussed                & Main typically required resources \\
\hline
Logical-based inferencing           & Parser producing logical meaning representations, inferencing engine,\\
                                    & resources to extract meaning postulates and common sense knowledge from.\\
Vector space semantic models        & Large monolingual corpus, possibly parser. \\
Surface string similarity measures  & Only preprocessing tools, e.g., \pos tagger, named-entity recognizer, which\\
                                    & are also required by most other methods.\\
Syntactic similarity measures       & Parser. \\
Similarity measures operating on    & Lexical semantic resources, possibly parser and/or semantic role labeling\\
symbolic meaning representations    & to produce semantic representations.\\
Machine learning algorithms         & Training/testing datasets, components/resources needed to compute features.\\
Decoding (transformation sequences) & Synonyms, hypernyms-hyponyms, paraphrasing/\textsc{te} rules.\\
Word/sentence alignment             & Large parallel or comparable corpora (monolingual or multilingual), possibly\\
                                    & parser.\\
Pivot language(s)                   & Multilingual parallel corpora.\\
Bootstrapping                       & Large monolingual corpus, recognizer. \\
Distributional hypothesis           & Monolingual corpus (possibly parallel or comparable).\\
Synchronous grammar rules           & Monolingual parallel corpus.\\
\hline
\end{tabular}
} 
\end{center}
\caption{Main ideas discussed and main resources they typically require.}
\label{ideas_resources}
\end{table}

\section*{Acknowledgments}

We thank the three anonymous reviewers for their valuable comments.
This work was funded by the Greek \textsc{pened} 
project ``Combined research in the areas of information retrieval, natural language processing, and user modeling aiming at the development of advanced search engines for document collections'', which was co-funded by the European Union (80\%) and the Greek General Secretariat for Research and Technology (20\%). 

\appendix
\section{On-line Resources Mentioned} \label{resources}

\subsection{Bibliographic Resources, Portals, Tutorials}

\begin{description}
\item[ACL 2007 tutorial on textual entailment:] {http://www.cs.biu.ac.il/$\sim$dagan\\
/TE-Tutorial-ACL07.ppt}.
\item[ACL Anthology:] {http://www.aclweb.org/anthology/}.
\item[Textual Entailment Portal:] {http://www.aclweb.org/aclwiki/index.php?\\
title=Textual\_Entailment\_Portal}.
\end{description}

\subsection{Corpora, Challenges, and their Datasets}

\begin{description}
\item[Cohn et al.'s paraphrase corpus:] Word-aligned paraphrases;\\  {http://www.dcs.shef.ac.uk/$\sim$tcohn/paraphrase\_corpus.html}.
\item[FATE:] The RTE-2 dataset with FrameNet annotations;\\
{http://www.coli.uni-saarland.de/projects/salsa/fate/}.
\item[MSR Paraphrase Corpus:] Paraphrase recognition benchmark dataset;\\ {http://research.microsoft.com/en-us/downloads/607d14d9-20cd-47e3-85bc-a2f65cd28042/}.
\item[Multiple-Translation Chinese Corpus:] Multiple English translations of Chinese news articles;\\ 
{http://www.ldc.upenn.edu/Catalog/CatalogEntry.jsp?catalogId=LDC2002T01}.
\item[RTE challenges, PASCAL Network of Excellence:] Textual entailment recognition challenges and their datasets; {http://pascallin.ecs.soton.ac.uk/Challenges/}.
\item[RTE track of NIST's Text Analysis Conference:] Continuation of \textsc{pascal}'s \rte;\\ 
{http://www.nist.gov/tac/tracks/}.
\item[Written News Compression Corpus:] Sentence compression corpus;\\
{http://jamesclarke.net/research/}.
\end{description}

\subsection{Implementations of Machine Learning Algorithms}

\begin{description}
\item[LIBSVM:] \svm implementation; {http://www.csie.ntu.edu.tw/$\sim$cjlin/libsvm/}.
\item[Stanford's Maximum Entropy classifier:] {http://nlp.stanford.edu/software/index.shtml}.
\item[SVM-Light:] \svm implementation; {http://svmlight.joachims.org/}.
\item[Weka:] Includes implementations of many machine learning algorithms;\\
{http://www.cs.waikato.ac.nz/ml/weka/}. 
\end{description}

\subsection{Implementations of Similarity Measures}

\begin{description}
\item[EDITS:] Suite to recognize textual entailment by computing edit distances; {http://edits.fbk.eu/}.
\item[WordNet::Similarity:] Implementations of WordNet-based similarity measures;\\ 
{http://wn-similarity.sourceforge.net/}.
\end{description}

\subsection{Parsers, POS Taggers, Named Entity Recognizers, Stemmers}

\begin{description}
\item[Brill's POS tagger:] {http://en.wikipedia.org/wiki/Brill\_tagger}.
\item[Charniak's parser:] {http://flake.cs.uiuc.edu/$\sim$cogcomp/srl/CharniakServer.tgz}.
\item[Collin's parser:] {http://people.csail.mit.edu/mcollins/code.html}.
\item[Link Grammar Parser:] {http://www.abisource.com/projects/link-grammar/}.
\item[MaltParser:] {http://w3.msi.vxu. se/$\sim$nivre/research/MaltParser.html}.
\item[MINIPAR:] {http://www.cs.ualberta.ca/$\sim$lindek/minipar.htm}.
\item[Porter's stemmer:] {http://tartarus.org/$\sim$martin/PorterStemmer/}.
\item[Stanford's named-entity recognizer, parser, tagger:] {http://nlp.stanford.edu/software/index.shtml}.
\end{description} 

\subsection{Statistical Machine Translation Tools and Resources}

\begin{description}
\item[Giza++:] Often used to train \ibm models and align words; {http://www.fjoch.com/GIZA++.html}.
\item[Koehn's Statistical Machine Translation site:] Pointers to commonly used \smt tools, resources; {http://www.statmt.org/}.
\item[Moses:] Frequently used \smt system that includes decoding facilities;\\
{http://www.statmt.org/moses/}.
\item[SRILM:] Commonly used to create language models;\\
{http://www.speech.sri.com/projects/srilm/}.
\end{description}

\subsection{Lexical Resources, Paraphrasing and Textual Entailment Rules}

\begin{description}
\item[Callison-Burch's paraphrase rules:] Paraphrase rules extracted from multilingual parallel corpora via pivot language(s); the implementation of the method used is also available;\\
{http://cs.jhu.edu/$\sim$ccb/}.
\item[DIRT rules:] Template pairs produced by \dirt; {http://demo.patrickpantel.com/}.
\item[Extended WordNet:] Includes meaning representations extracted from WordNet's glosses;\\ {http://wordnet.princeton.edu/}.
\item[FrameNet:] {http://framenet.icsi.berkeley.edu/}.
\item[Nomlex:] English nominalizations of verbs; {http://nlp.cs.nyu.edu/nomlex/}
\item[TEASE rules:] Textual entailment rules produced by \tease;\\ {http://www.cs.biu.ac.il/$\sim$szpekti/TEASE\_collection.zip}.
\item[VerbNet:] {http://verbs.colorado.edu/$\sim$mpalmer/projects/verbnet.html}.
\item[WordNet:] {http://xwn.hlt.utdallas.edu/}.
\item[Zhao et al.'s paraphrase rules:] Paraphrase rules with slots corresponding to \pos tags, extracted from multilingual parallel corpora via pivot language(s);\\
{http://ir.hit.edu.cn/phpwebsite/index.php?\\
module=documents\&JAS\_DocumentManager\_op=viewDocument\&JAS\_Document\_id=268}.
\end{description}

\bibliographystyle{theapa}
\bibliography{paraphrasing_textual_entailment_survey}
\end{document}